\RequirePackage{silence}
\PassOptionsToPackage{svgnames,dvipsnames,table}{xcolor}
\PassOptionsToPackage{capitalize}{cleveref}
\documentclass[table]{ai2style/ai2}
\usepackage{lmodern}
\usepackage{fix-cm}
\usepackage{amssymb}
\usepackage{bigdelim}
\usepackage{todonotes}
\usepackage{longtable}
\usepackage{float}
\usepackage{tabularx}
\usepackage{tabularray}
\usepackage[most]{tcolorbox} 
\usepackage{svg}
\usepackage[absolute]{textpos}
\usepackage{fdsymbol}
\usepackage[utf8]{inputenc}
\usepackage[T1]{fontenc} 
\usepackage{url}
\usepackage{booktabs}    
\usepackage{amsfonts}
\usepackage{nicefrac}    
\usepackage{microtype}   
\usepackage[table]{xcolor}
\usepackage{amsmath}
\DeclareMathSizes{11}{11}{8}{6}
\usepackage{csquotes}

\usepackage{siunitx}
\usepackage{graphicx}
\usepackage{arydshln}
\usepackage{wrapfig}
\usepackage{enumitem}
\usepackage{soul} %

\usepackage{multirow}
\usepackage{xspace}
\usepackage{adjustbox}
\usepackage{pifont}
\usepackage{makecell}
\usepackage{bold-extra}

\usepackage{caption}

\usepackage{hyperref}
\definecolor{linkcolor}{RGB}{0, 0, 128}
\hypersetup{
     colorlinks   = true,
     citecolor    = linkcolor,
     linkcolor    = linkcolor,
     urlcolor     = linkcolor,
}
\newcommand{\loopvlappendixtocstyle}{%
  \setlength{\cftsecindent}{\cftsubsecindent}%
  \setlength{\cftsecnumwidth}{\cftsubsecnumwidth}%
  \setlength{\cftbeforesecskip}{\cftbeforesubsecskip}%
  \renewcommand{\cftsecdotsep}{\cftsubsecdotsep}%
  \renewcommand{\cftsecleader}{\cftsubsecleader}%
}
\usepackage{listings}

\setlist[itemize]{leftmargin=*,itemsep=0em,parsep=0.3em,topsep=0.3em}

\definecolor{iclrdeepblue}{RGB}{0,0,128}
\definecolor{maroon}{HTML}{F26035}
\definecolor{yellow}{HTML}{FDBC42}
\definecolor{lavender}{HTML}{734f96}
\definecolor{darkergrey}{HTML}{444444}
\definecolor{midgrey}{HTML}{e6eded}
\definecolor{ai2pink}{HTML}{5477C4}%
\definecolor{ai2midpink}{HTML}{fad3e5}
\definecolor{ai2lightpink}{HTML}{fbecf3}
\definecolor{ai2midwhite}{HTML}{f2e5d9}
\definecolor{ai2offwhite}{HTML}{fbf4ee}
\definecolor{ai2green}{HTML}{0fcb8c}
\definecolor{ai2lightgreen}{HTML}{e7f9f3}
\definecolor{ai2darkgreen}{HTML}{105257}
\definecolor{ai2purple}{HTML}{B932EB}
\definecolor{ai2lightpurple}{HTML}{f7e8fc}
\definecolor{neutralEight}{HTML}{343434}
\definecolor{neutralFive}{HTML}{838383}
\definecolor{neutralThree}{HTML}{bebebe}
\definecolor{neutralOne}{HTML}{dedede}
\definecolor{lightgrey}{HTML}{fafcfc}

\usepackage{tikz}
\usepackage[framemethod=TikZ]{mdframed}
\usepackage{xpatch}
\usetikzlibrary{shadows,arrows.meta,positioning,calc,shapes.geometric}
\usepackage{pgfplots}
\pgfplotsset{compat=1.18}

\mdfdefinestyle{mdpurplebox}{%
  roundcorner=10pt,
  linewidth=1pt,
  skipabove=12pt,
  skipbelow=12pt,
  innertopmargin=9pt,
  innerbottommargin=9pt,
  linecolor=black,
  nobreak=true,
  backgroundcolor=Orchid!10,
  shadow=true,
  shadowsize=6pt,
  shadowcolor=black!30,
  frametitleaboveskip=8pt,
  frametitlebelowskip=8pt,
  frametitlebackgroundcolor=Violet!50!black,
  frametitlefont=\bfseries\sffamily\color{white},
  frametitlerule=true,
}
\makeatletter
\xpatchcmd{\endmdframed}
  {\aftergroup\endmdf@trivlist\color@endgroup}
  {\endmdf@trivlist\color@endgroup\@doendpe}
  {}{}
\makeatother

\DeclareUnicodeCharacter{2011}{\nobreakdash-}
\DeclareUnicodeCharacter{2013}{--}
\DeclareUnicodeCharacter{2019}{\textquoteright}
\DeclareUnicodeCharacter{201C}{``}
\DeclareUnicodeCharacter{201D}{''}
\DeclareUnicodeCharacter{202F}{\nobreak\hspace{0.16667em}}
\DeclareUnicodeCharacter{21D2}{\ensuremath{\Rightarrow}}
\DeclareUnicodeCharacter{220E}{\ensuremath{\square}}

\definecolor{darkred}{RGB}{156, 39, 33}
\definecolor{darkblue}{RGB}{31, 90, 153}
\definecolor{forestgreen}{rgb}{0.13, 0.55, 0.13}
\definecolor{olmoDarkBlue}{HTML}{012e59}
\definecolor{olmoBlue}{HTML}{265ed4}
\definecolor{olmoLightBlue}{HTML}{012e59}
\definecolor{olmoTeal}{HTML}{00d5ff}
\definecolor{olmoYellow}{HTML}{ffbb00}
\definecolor{olmoOrange}{HTML}{ff9100}

\newcommand{\huggingface}{\raisebox{-1.5pt}{\includegraphics[height=1.05em]{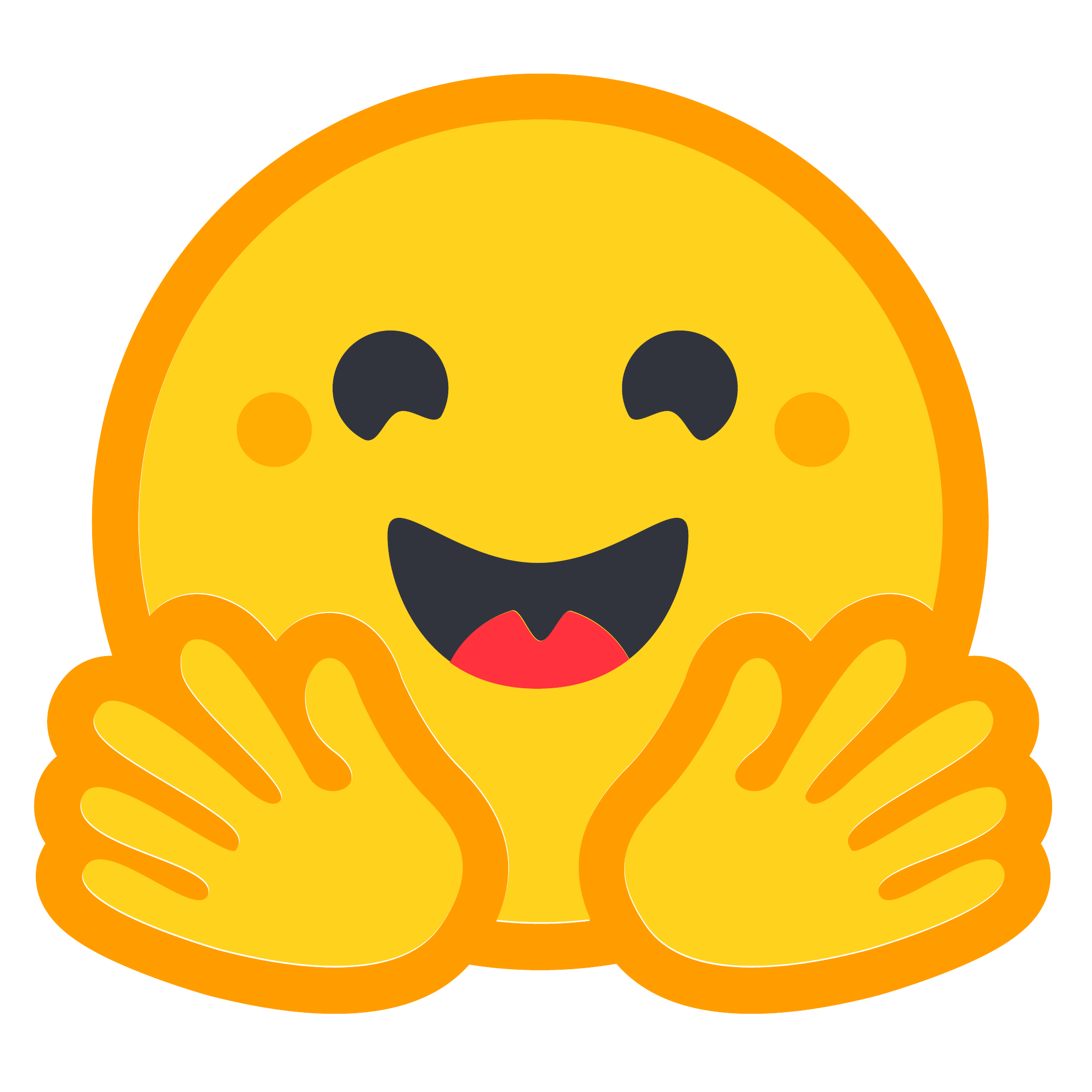}}\xspace}
\newcommand{\modelscope}{\raisebox{-1.5pt}{\includegraphics[height=1.05em]{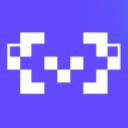}}\xspace}
\newcommand{\projectlink}[2]{\href{#1}{\textcolor{ai2accent}{\texttt{#2}}}}

\newcommand{\github}{\raisebox{-1.5pt}{\includegraphics[height=1.05em]{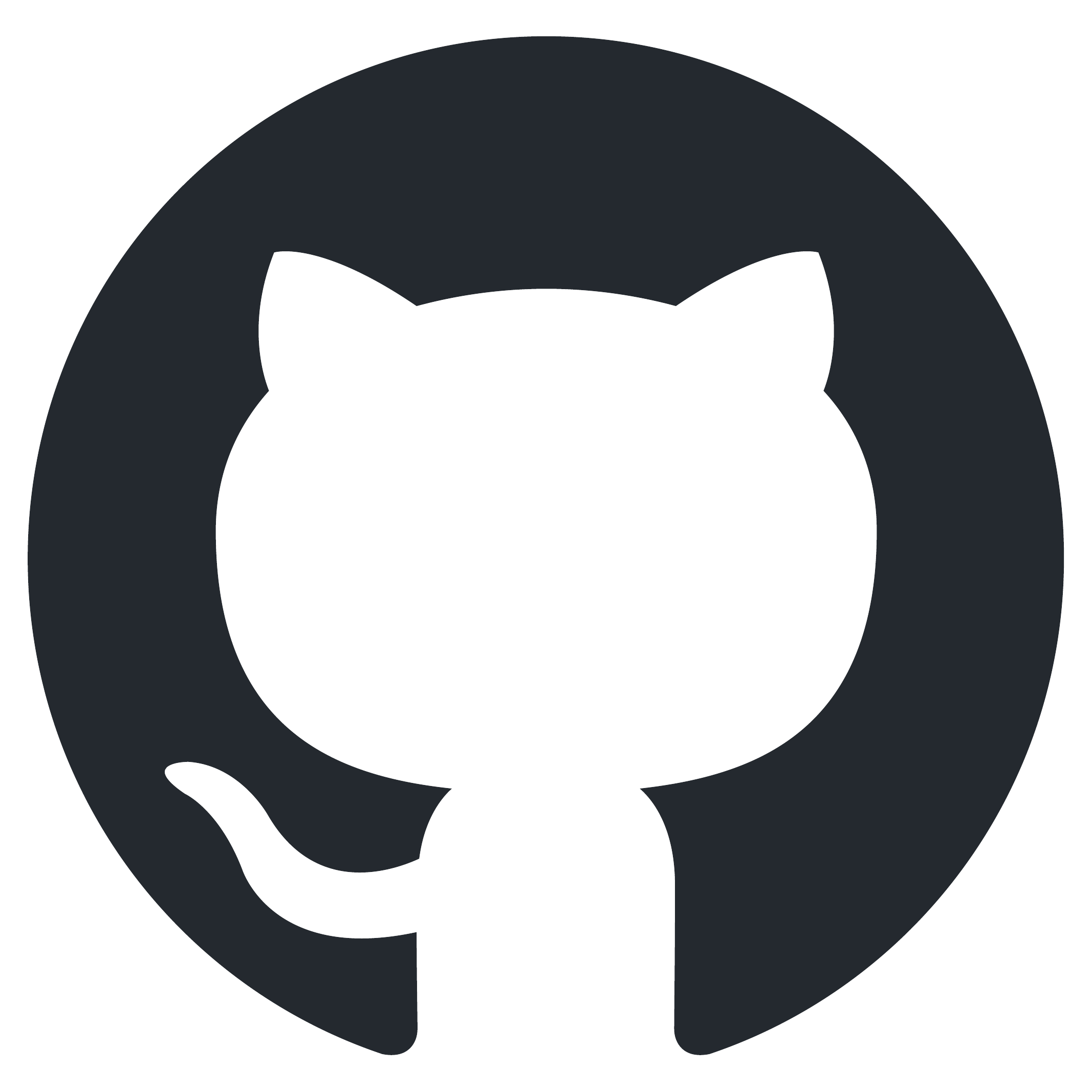}}\xspace}

\usepackage{setspace}

\usepackage{nicematrix}
\newcolumntype{L}[1]{>{\raggedright\let\newline\\\arraybackslash\hspace{0pt}}m{#1}}
\newcolumntype{C}[1]{>{\centering\let\newline\\\arraybackslash\hspace{0pt}}m{#1}}
\newcolumntype{R}[1]{>{\raggedleft\let\newline\\\arraybackslash\hspace{0pt}}m{#1}}
\newcolumntype{P}[1]{>{\centering\let\newline\\\arraybackslash\columncolor{ai2lightpink}}m{#1}}

\newcommand{\LoopVLTrainingTokens}{0.14T}

\DeclareRobustCommand{\LVFig}[1]{}
\DeclareRobustCommand{\LVCitationPending}[1]{}
\newcommand{\LVAuthorCheck}[2]{}
\newenvironment{LVQuestion}{%
  \begin{tcolorbox}[colback=ai2offwhite,colframe=ai2pink,boxrule=0.5pt,arc=3pt,left=6pt,right=6pt,top=5pt,bottom=5pt]
  \centering\itshape
}{\end{tcolorbox}}

\newcommand{\LVAssetCaption}[1]{\csname LVCaption#1\endcsname}
\expandafter\def\csname LVCaptionP01\endcsname{Visual overview of cross-cycle reading behavior. The rows illustrate color, location, counting, and text-recognition examples. The left group compares the eighth layer of the first L invocation across model cycles (zero-based executed-layer indices 7 and 71); the right group compares the eighth H layer (indices 55 and 119). Each group includes cycle-specific received-attention maps and their differences. The same physical parameters are compared within each group. These qualitative examples introduce the image-space view of recurrence; they are neither strict cycle-end comparisons nor estimates of the phenomenon's prevalence.}
\expandafter\def\csname LVCaptionP02\endcsname{Spatial attention concentration over unfolded computation. The entropy trace shows variation across the two model cycles, with lower values in much of the later computation rather than a monotonic decrease at every layer. The visual panels compare the eighth and final H layers in both cycles, at zero-based indices 55, 63, 119, and 127, using a common received-attention scale. The experiment records a summary entropy reduction of 0.323. Its aggregation should be distinguished from the strict endpoint statistics reported separately.}
\expandafter\def\csname LVCaptionP03\endcsname{Strict cycle-end comparison at the final layer of the same H module. Thin lines represent individual analyzed samples and the highlighted line represents their mean. Across the displayed set, spatial entropy decreases from 0.907 to 0.553, the visual-token fraction required to cover 80\% of attention decreases from 0.418 to 0.083, and Gini concentration increases from 0.405 to 0.756. The paired comparison controls the physical module and local layer. Consistency across the displayed samples describes this diagnostic set, not every benchmark example.}
\expandafter\def\csname LVCaptionP04\endcsname{Visual-attention mass under aligned L-module reuse. Rows index the 16 physical layers of the shared L module, and columns index its three invocation slots within a model cycle. The first two panels report the corresponding first- and second-cycle values; the third reports their difference. The recorded mean change is \ensuremath{-}0.219. The comparison demonstrates that the same physical layers at corresponding invocation slots need not allocate the same amount of attention to vision after the recurrent state has changed.}
\expandafter\def\csname LVCaptionP05\endcsname{Query/key compatibility and visual-value diagnostics. The matrix combines early and late queries with early and late visual keys; visual-attention mass is lower with late keys for both query choices (0.417 and 0.410, versus 0.614 and 0.645 with early keys). The companion panels show recorded loss changes for answer-side and instruction-side visual-value removal probes, with means of 0.618 and 3.493. These probes distinguish attention selection from value transmission within their intervention settings. They do not establish that keys are the sole determinant of recurrent visual behavior.}
\expandafter\def\csname LVCaptionP06\endcsname{Linear recoverability of the visual reference across recurrent calls. The plot reports the probe error as 1\ensuremath{-}R\textsuperscript{2}, beginning after projection and continuing through the L/H invocations. The recorded recovery score changes from 0.613 after projection to 0.310 at the recurrent endpoint, corresponding to errors of 0.387 and 0.690. Intermediate variation is nonmonotonic. Lower linear recoverability concerns the selected probe and reference representation; it does not establish that visual information has disappeared from the model.}
\expandafter\def\csname LVCaptionP07\endcsname{Cross-loop rank reallocation at the final H layer. For the selected color and counting examples, target-region attention rises from 8.8\% to 55.9\% and from 6.7\% to 20.5\%, respectively. The recorded rank correlations are 0.474 and 0.350. Among second-cycle top-quartile positions, 23.6\% and 31.9\% were in the first-cycle bottom half. Both comparisons align the same physical H layer at strict cycle endpoints. The cases demonstrate that later attention need not preserve the earlier ranking; their selection does not estimate how frequently this occurs.}
\expandafter\def\csname LVCaptionP08\endcsname{Qualitative evidence reallocation in two RealWorldQA images. Each row shows the source image, first- and second-cycle attention maps, and their difference. In the road example, later attention is concentrated near debris beside the white van. In the night-market example, attention shifts toward the relevant left-side stall and paper-container region. The examples make changes in question-related spatial reading visible, but do not on their own establish a transition from an incorrect first-cycle answer to a correct second-cycle answer.}
\expandafter\def\csname LVCaptionP09\endcsname{State-conditioned reading behavior of a shared attention head. The visual panels compare the head identified in the record as H-module layer 7, head 8, across the two model cycles. In the selected counting example, target-region enrichment changes from 0.05 to 13.02 and spatial entropy from 0.677 to 0.007. The paired plot places the example alongside the analyzed sample set. Since the parameters are shared, the comparison illustrates different reading behavior under different recurrent states, rather than different first- and second-cycle parameter sets.}
\expandafter\def\csname LVCaptionP10\endcsname{Attention concentration at the model-cycle boundary. The Gini trace is indexed by zero-based executed-layer calls. Concentration changes from 0.405 at index 63, the end of the first H invocation, to 0.858 at index 64, the start of the second cycle's first L invocation. This aligns an attention change with the recurrent boundary. The comparison changes both the recurrent state and module type, so it does not isolate the cause of the concentration increase.}
\expandafter\def\csname LVCaptionP12\endcsname{State updates across module boundaries on 32 multimodal samples: 16 VMCBench and 16 AI2D examples. Panels (a,b) analyze all multimodal positions, and panels (c,d) analyze visual positions. The reported mean update statistics are 56.39 and 56.54, with second-half/first-half ratios of 0.97 and 0.98, respectively. Similarity matrices compare states after 16, 32, \ldots{}, 128 executed layers. These diagnostics show continued state changes in the second cycle; an absolute update magnitude alone is not evidence that those changes improve the answer.}
\expandafter\def\csname LVCaptionP13\endcsname{Alignment of intermediate readouts with the final model distribution on the same 32-sample collection. Background bands mark complete 16-layer module invocations. The recorded mean KL distance is 9.18 after the first model cycle, 7.27 after the third L invocation of the second cycle, and 3.01 at the eighth layer of the final H module. The final value is zero because the reference distribution is compared with itself. The informative feature is the late alignment trajectory, not that identity; the reference is not a ground-truth answer distribution.}

\title{LoopVL: Recurrent Visual Intelligence}

\authorThree{Zhe Qian\textsuperscript{1,2,*}}
\authorThree{Ziyang Gong\textsuperscript{3,*}}
\authorThree{Zhongxing Xu\textsuperscript{4,*}}
\authorThree{Hehan Li\textsuperscript{2,*\textdagger}}
\authorThree{Zhonghua Wang\textsuperscript{4}}

\authorFour{Hanmeng Liu\textsuperscript{2}}
\authorFour{Liwei Qian\textsuperscript{2}}
\authorFour{Xin Pei\textsuperscript{2}}
\authorFour{Peizhi Xu\textsuperscript{2}}
\authorFour{Fei Luo\textsuperscript{7}}
\authorFour{Mingxuan Wang\textsuperscript{7}}

\authorFive{Xue Yang\textsuperscript{3}}
\authorFive{Shiwei Liu\textsuperscript{5,6}}
\authorFive{Yanbiao Ma\textsuperscript{1,\textdaggerdbl}}
\authorFive{Junchi Yan\textsuperscript{3,\textdaggerdbl}}
\authorFive{Jungong Han\textsuperscript{8,\textdaggerdbl}}

\DeclareRobustCommand{\titleinstitutionlogos}{%
  \makebox[\linewidth][r]{%
    \raisebox{-.5\height}{\includegraphics[trim=3.84bp 4.32bp 7.68bp 4.8bp,clip,height=5.85mm]{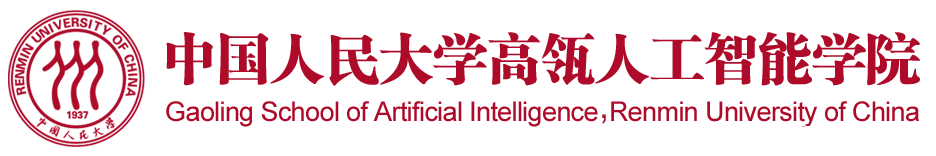}}%
    \hspace{2mm}%
    \raisebox{-.5\height}{\includegraphics[trim=4bp 2.864bp 4.339bp 4.375bp,clip,height=5.85mm]{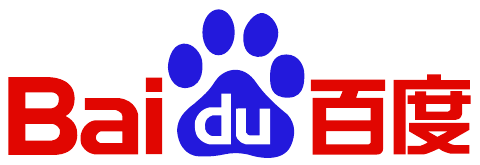}}%
    \hspace{2mm}%
    \raisebox{-.5\height}{\includegraphics[height=8.28mm]{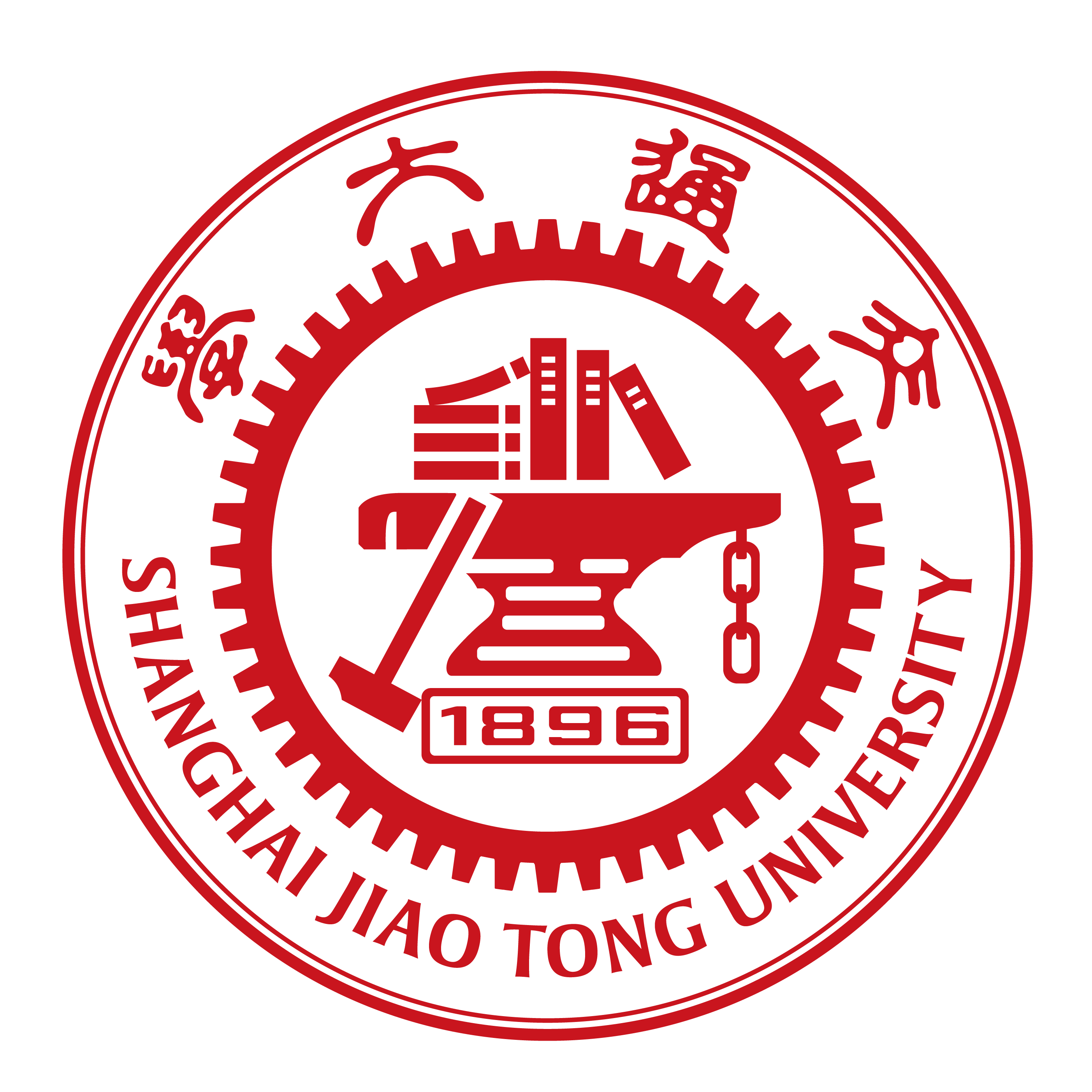}}%
    \hspace{2mm}%
    \raisebox{-.5\height}{\includegraphics[trim=0bp 0bp 746bp 0bp,clip,height=7.56mm]{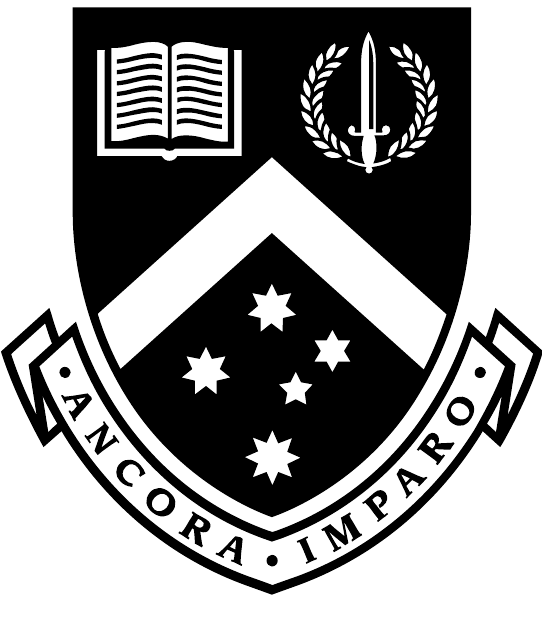}}%
    \hspace{2mm}%
    \raisebox{-.5\height}{\includegraphics[trim=126bp 85bp 123bp 107bp,clip,height=8.28mm]{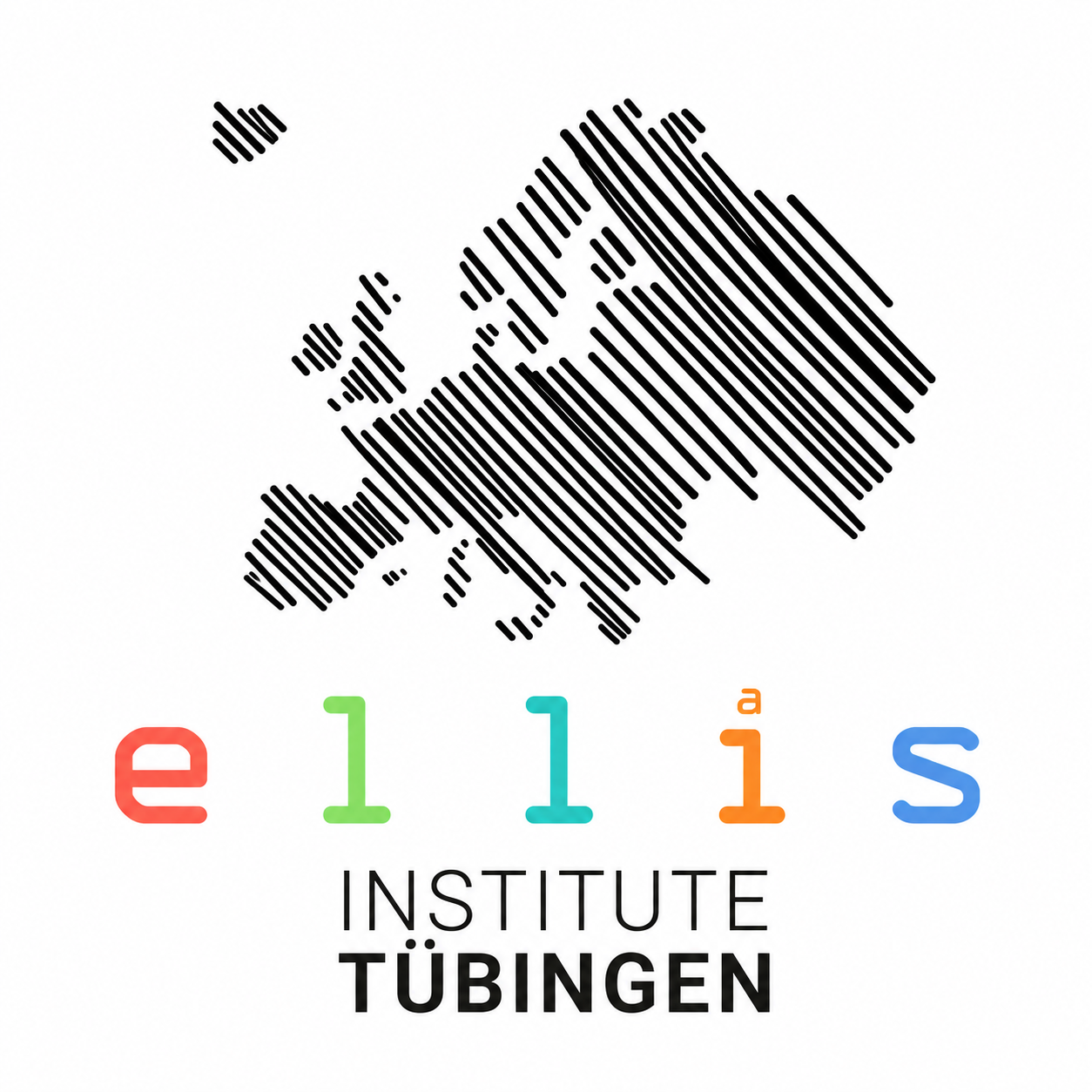}}%
    \hspace{2mm}%
    \raisebox{-.5\height}{\includegraphics[height=8.28mm]{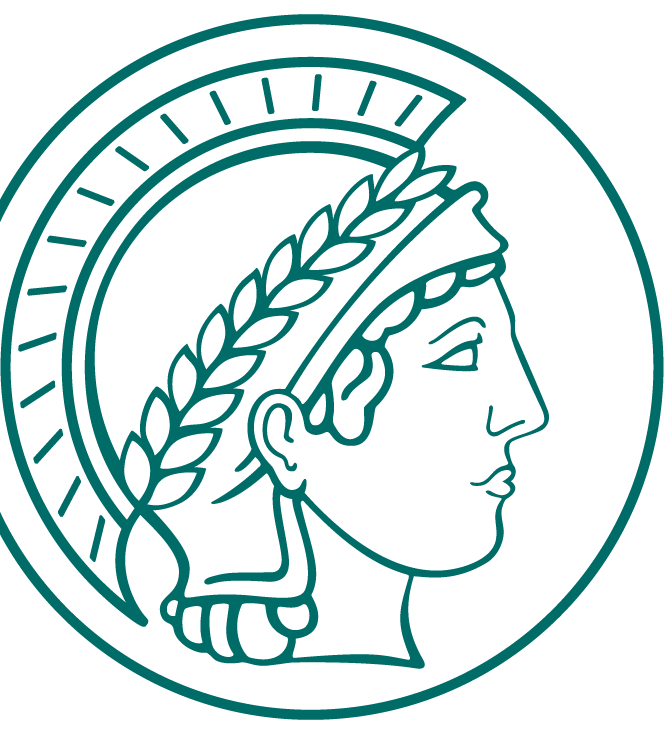}}%
    \hspace{2mm}%
    \raisebox{-.5\height}{\includegraphics[height=8.28mm]{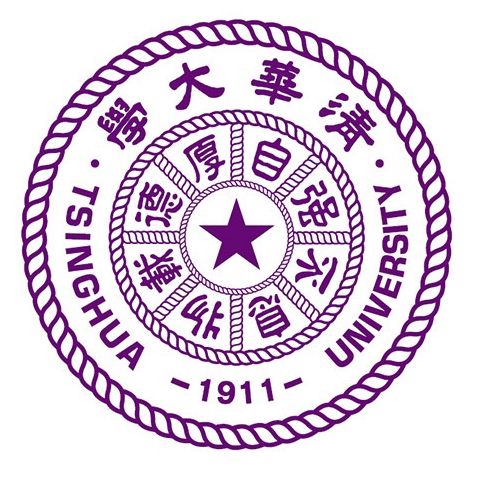}}%
    \hspace{2mm}%
    \raisebox{-.5\height}{\includegraphics[height=5.85mm]{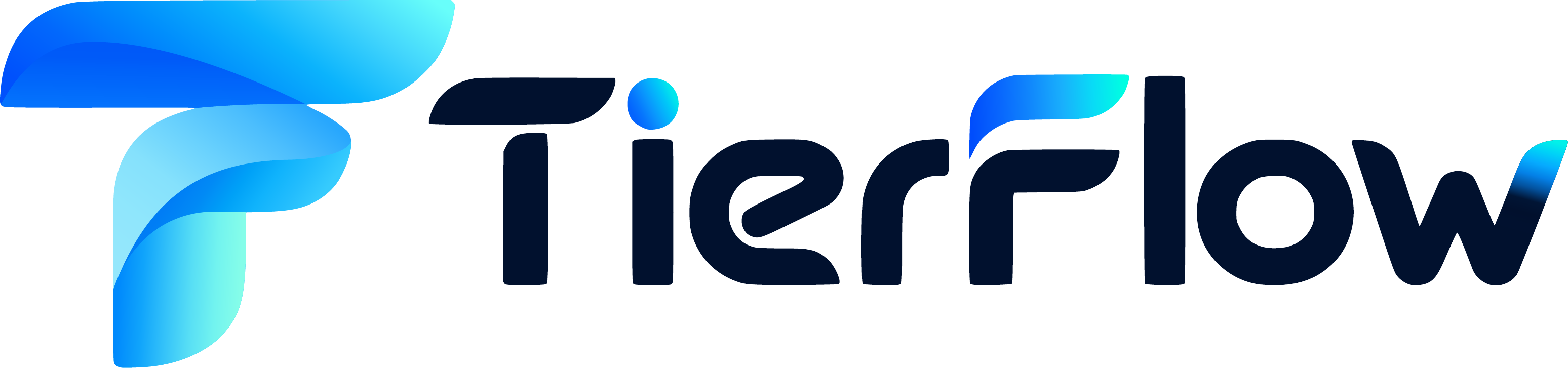}}%
  }%
}
\let\papertitletext\titlelist
\renewcommand{\titlelist}{%
  \begin{minipage}[t]{\linewidth}
    \vspace*{-9.7mm}
    \centering\titleinstitutionlogos\par
    \vspace{1.6mm}
    {\color{ai2pink!25!white}\hrule height 0.35pt}
    \vspace{6.8mm}
    \raggedright\adjustbox{max width=\linewidth}{\papertitletext}\par
  \end{minipage}%
}

\DeclareRobustCommand{\institutionnames}{%
  \begin{minipage}[t]{\linewidth}
    \small
    \textsuperscript{1}Gaoling School of Artificial Intelligence, Renmin University of China\quad
    \textsuperscript{2}Baidu\par
    \vspace{1mm}
    \textsuperscript{3}Shanghai Jiao Tong University\quad
    \textsuperscript{4}Monash University\quad
    \textsuperscript{5}ELLIS Institute Tübingen\par
    \vspace{1mm}
    \textsuperscript{6}Max Planck Institute for Intelligent Systems\quad
    \textsuperscript{7}TierFlow Team\quad
    \textsuperscript{8}Tsinghua University\par
    \vspace{1mm}
    \footnotesize
    \textsuperscript{*}Equal contribution.\quad
    \textsuperscript{\textdagger}Project leader.\quad
    \textsuperscript{\textdaggerdbl}Corresponding author.
  \end{minipage}%
}
\affiliation{\institutionnames}

\abstract{
We introduce \textbf{LoopVL} to study whether Loop Transformers can be effectively extended to vision-language models. LoopVL combines Module-Loop and Model-Loop computation to iteratively update a unified vision-language state through shared modules. We train LoopVL from scratch through language pre-training, multimodal training, and post-training. LoopVL outperforms a range of similarly sized and larger non-recurrent models on multimodal understanding and visual reasoning benchmarks. We also observe Visual Aha Moments in LoopVL, characterized by pronounced shifts in visual attention across loops. LoopVL provides practical evidence for recurrent vision-language modeling and offers an intuitive perspective on how shared parameters can support deeper multimodal computation over continuously evolving visual-language states.

}

\metadata[\vspace{-4mm}\hspace{0.5mm}\github Code:]{\projectlink{https://github.com/Tier-Flow/LoopVL}{LoopVL}}

\metadata[\vspace{.5em}\hspace{0.5mm}\huggingface Models:]{%
  \projectlink{https://huggingface.co/TierFlow/LoopVL}{LoopVL-1B}%
  \hspace{2em}%
  {\sffamily\bfseries\modelscope Models:}\enspace
  \projectlink{https://www.modelscope.cn/models/Eternity123/LoopVL}{LoopVL-1B}%
}

\begin{document}
\maketitle

\enlargethispage{60pt}
\par\vspace{-8pt}
\noindent\begin{minipage}{\textwidth}
  \centering
  \includegraphics[width=1.0\linewidth]{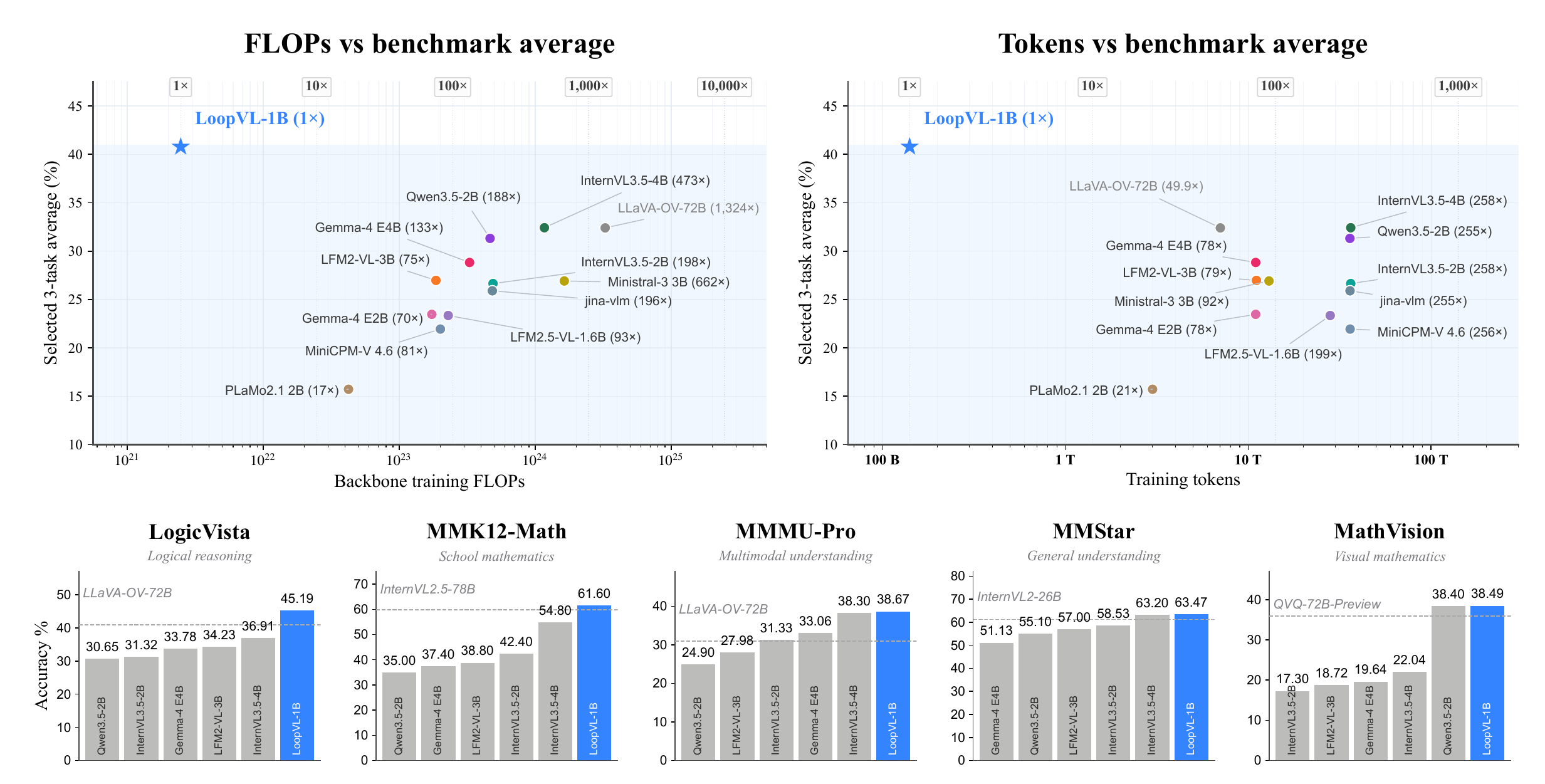}
  \captionsetup{format=plain,labelsep=colon,font=small,labelfont=normalfont,
    justification=justified,singlelinecheck=false,skip=4pt}
  \captionof{figure}{{\bfseries Compute and data efficiency.} Top: Backbone FLOPs and training tokens versus mean accuracy on LogicVista, MMMU-Pro, and MathVision; ratios are reported relative to LoopVL. Bottom: Five benchmark comparisons. Blue denotes LoopVL; dashed lines indicate selected published large-model scores.}
  \label{fig:lv-compute-benchmark-overview}
\end{minipage}\par

\newpage
\setcounter{tocdepth}{2}
\tableofcontents
\newpage

\section{Introduction}
\label{sec:lv-1}

Recurrent Transformers were first introduced in the form of the Universal Transformer \citep{dehghani2018universal} and have recently re-emerged as an important alternative to conventional depth scaling. Instead of stacking independently parameterized Transformer layers, recurrent architectures increase effective computational depth by repeatedly applying shared modules, enabling deeper computation without a proportional increase in parameter count \citep{csordas2024moeut}. Subsequent recurrent-depth language models \citep{geiping2025scalinglatent,zhu2025scalinglatent} and hierarchical reasoning models \citep{wang2025hierarchical,wang2026hrmtext} have further developed this idea, showing that shared parameters can perform different computations as the underlying hidden states evolve over successive recurrent steps.

A growing body of work suggests that recurrent computation is particularly well suited to tasks requiring iterative state updates. Prior studies have shown that repeated computation can support in-context learning, algorithmic learning, and complex reasoning, while allowing relatively shallow shared-parameter models to approach the performance of substantially deeper non-shared architectures on certain tasks \citep{giannou2023looped,yang2023looped,gatmiry2024can,saunshi2025latentthoughts}. More recently, recurrent architectures have begun to scale to billion-parameter language models \citep{gao2026loopies}, with increasing attention to the effects of recurrence depth, hierarchical organization, and additional test-time computation.

However, existing studies have largely focused on language models, while extending recurrent computation to vision--language models introduces a new challenge. Images provide candidate evidence with explicit spatial structure \citep{wang2024qwen2vl}, whereas the instruction determines which regions are relevant to the current query. After one round of vision--language interaction, both the representations and the relative importance of visual tokens may change. Subsequent recurrent steps may therefore continue attending to the same regions or reallocate attention toward previously overlooked areas. Thus, in vision--language models, recurrent computation is not merely a means of increasing computational depth; it must also operate over continuously evolving vision--language states. This leads to our central question:

\begin{LVQuestion}
Can Loop Transformers be effectively extended to vision--language models?
\end{LVQuestion}

Moreover, the visual modality provides a natural observation window for understanding recurrent computation itself. Because visual tokens preserve a spatial correspondence with image patches, we can directly compare which regions the same physical layers attend to across different recurrent steps and observe whether initially low-importance regions become prominent in later computation. This makes it possible to map recurrent state transitions back to concrete regions in the input image.

Motivated by these observations, we introduce LoopVL, which extends recurrent computation to vision--language modeling. The core idea is to perform persistent recurrent computation over a unified multimodal state, allowing visual information and language instructions to interact continuously across multiple stages of computation. LoopVL repeatedly updates the joint vision--language state using shared parameters, enabling deeper multimodal computation without introducing additional parameters. This design retains the parameter efficiency of recurrent Transformers while allowing additional computation to act directly on evolving vision--language representations. We further study how visual information changes throughout the recurrent process and find that the model reorganizes and reuses visual evidence during later stages of computation.

Our contributions are threefold:

\begin{itemize}
    \item \textbf{Recurrent vision--language modeling.} We introduce LoopVL and systematically investigate the feasibility of extending recurrent Transformers to vision--language models, demonstrating that repeated computation with shared parameters can be effectively applied to multimodal modeling.
    \item \textbf{Comprehensive validation of multimodal recurrence.} We perform vision–language pretraining and post-training of LoopVL and evaluate it across a range of multimodal tasks and recurrent configurations, providing systematic empirical evidence for scaling recurrent Transformers to vision–language models.
    \item \textbf{Analysis of recurrent visual dynamics.} We systematically study how visual information evolves throughout recurrent computation, revealing that later recurrent steps do not merely repeat earlier computation, but continuously reorganize and reuse visual evidence.
\end{itemize}

\section{The LoopVL Model}
\label{sec:lv-2}

This section focuses on how LoopVL organizes recurrent computation and how visual information continues to evolve throughout this process. We first introduce the overall architecture and the coupled L/H backbone, and then explain why visual tokens should be continuously updated across recurrent invocations. Next, we motivate the hierarchical recurrent design and distinguish Model-Loop from Module-Loop, showing how the two jointly form LoopVL's nested recurrent execution pattern over a continuously evolving visual--language state.

\subsection{Architecture}
\label{sec:lv-2-1}
\begin{figure}[!t]
\centering
\includegraphics[width=\linewidth]{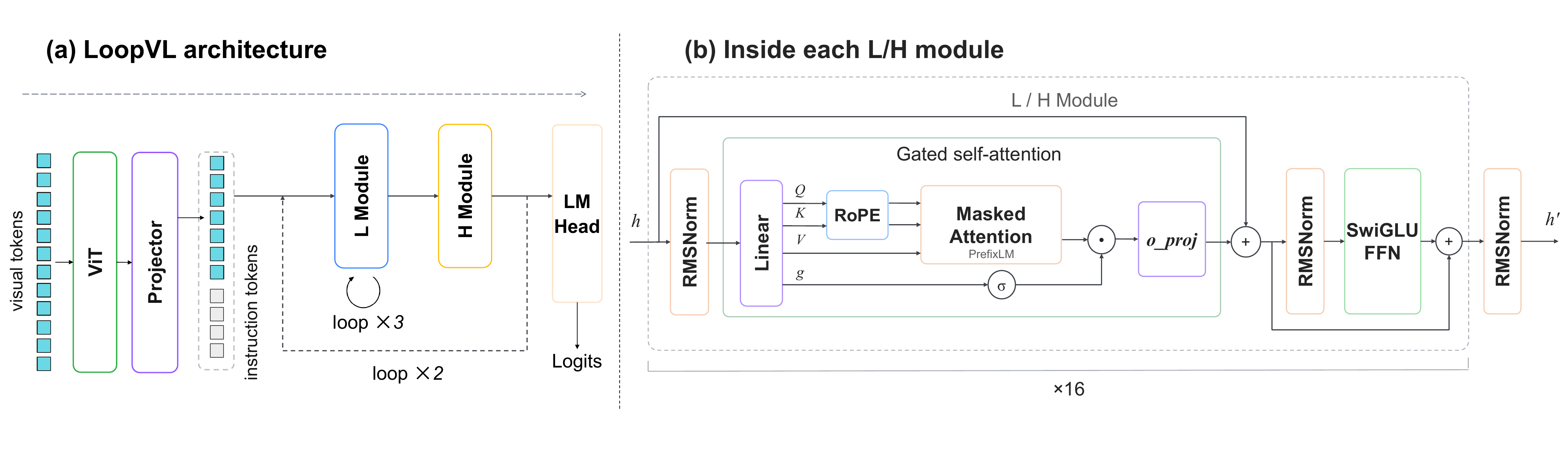}
\caption{LoopVL architecture and the structure of its recurrent modules. (a) A visual encoder and projector supply language-space visual positions that enter the recurrent backbone alongside instruction tokens. Module-Loop repeats the complete L stack, and Model-Loop repeats the coupled L/H update cycle; the default schedule is [L\ensuremath{\times}3\ensuremath{\rightarrow}H]\ensuremath{\times}2. (b) Each L/H module contains 16 Transformer layers with normalization, gated self-attention, and a SwiGLU feed-forward path. L and H have separate parameters, each reused across its invocations.}
\label{fig:lv-architecture}
\end{figure}

LoopVL uses the Penguin Vision Encoder as its frozen visual encoder, together with a trainable visual interface and our own pretrained recurrent language backbone. The language backbone is built upon HRM-Text's open-source framework. Its pretraining data are based on the publicly released HRM-Text training data, augmented with additional data curated by us. The Penguin Vision Encoder produces a 1024-dimensional feature for each retained image patch. A two-layer GELU projector maps each visual feature from 1024 to 1536 dimensions through a 1536-dimensional hidden layer. The projected embeddings replace the reserved visual slots in a sequence containing the conditioning header, image, instruction, and response. Figure~\ref{fig:lv-architecture} illustrates both the overall multimodal computation and the Transformer structure within each L/H module. \citep{wang2026hrmtext}

The language backbone contains separate L and H parameter stacks, each consisting of 16 Transformer layers. Each stack shares its own parameters across repeated invocations, while L and H do not share parameters with each other. Under the default H2L3 configuration, the execution order is

\begin{equation}
\underbrace{L\rightarrow L\rightarrow L\rightarrow H}_{\text{model loop 1}}
\rightarrow
\underbrace{L\rightarrow L\rightarrow L\rightarrow H}_{\text{model loop 2}}.
\label{eq:lv-recurrent-order}
\end{equation}

An L or H symbol denotes a complete module invocation rather than an individual attention layer. Consequently, although the backbone stores only two 16-layer Transformer stacks, a complete forward execution performs 128 Transformer-layer calls.

Within each module, the Transformer blocks shown in Figure~\ref{fig:lv-architecture} consist of normalization, gated self-attention, residual paths, and a SwiGLU feed-forward transformation. Visual and text tokens use different rotary positional encodings. Visual tokens use two-dimensional spatial RoPE, with their row and column coordinates on the image grid encoding spatial position. Text tokens are also assigned two-dimensional coordinates, but the two coordinates advance synchronously along the text sequence, reducing to an effect equivalent to the original one-dimensional RoPE. The PrefixLM mask is applied within the attention computation, while the gate inside the attention sublayer modulates its output path. These operations are reused whenever the module is invoked again.

The recurrent computation maintains coupled L and H states. The H state is initialized from the scaled input embeddings, whereas the L state is initialized to zero and retained across model cycles. Within each cycle, L is repeatedly updated under the current high-level condition, followed by an H update that incorporates the refined L state. The final H state is passed to the language-modeling head. This differs from simply stacking a sequence of independent blocks: the shared L transformation is revisited multiple times before H updates the condition used by the next model cycle. At the start of each model cycle, LoopVL re-injects a fixed anchor from the initial projected visual embeddings into the visual positions of the H state. A query-conditioned token-wise visual gate and a learnable cycle-specific scale control the strength of this re-injection.

During training, we use selective backpropagation across recurrent invocations. For the default H2L3 schedule, the recurrent execution order is \(\mathrm{L1}\!\rightarrow\!\mathrm{L2}\!\rightarrow\!\mathrm{L3}\!\rightarrow\!\mathrm{H1}\!\rightarrow\!\mathrm{L4}\!\rightarrow\!\mathrm{L5}\!\rightarrow\!\mathrm{L6}\!\rightarrow\!\mathrm{H2}\). During backpropagation warmup, the active backward graph gradually expands from the final two invocations (L6 and H2) to the final five invocations (L4, L5, L6, H1, and H2). The earlier L calls (L1, L2, and L3) remain excluded from gradient computation. All recurrent invocations are still executed in the forward pass. All valid sequence positions participate in the module invocations, but their attention visibility differs. Header, image, and instruction positions jointly form a bidirectional prefix. Response positions can attend to this prefix and to preceding response positions, while prefix positions cannot attend to the response.

\begingroup
\clubpenalties 2 10000 0
\widowpenalties 2 10000 0
\subsection{Motivation}
\label{sec:lv-2-4}

LoopVL's recurrent architecture is inspired by biological systems. Biological systems do not perform all information processing at a single timescale; instead, they coordinate high-level information with fast local computation through state changes operating at different timescales. HRM and HRM-Text bring this idea into recurrent models by organizing computation through coupled high-level and low-level states: the low-level state is repeatedly updated at a faster timescale, while the high-level state changes its computational condition at a slower rate. LoopVL inherits this hierarchical recurrent structure and further investigates how such computation operates over a joint visual--language state.

Vision--language tasks naturally involve computation at different levels of granularity. A model must process fine-grained visual evidence such as color, text, boundaries, and local spatial relations, while also maintaining a high-level semantic state determined by the current question and the overall scene. Motivated by this observation, we adopt a nested L/H update scheme: under a relatively stable high-level state, the L module is invoked repeatedly to refine the current visual--language representation; the H module then integrates these updates and forms the high-level state for the next recurrent cycle. This organization allows local state refinement and global state updating to occur at different recurrent timescales.

This motivation also determines the respective roles of Module-Loop and Model-Loop. Module-Loop provides denser state updates under the same high-level condition, whereas Model-Loop allows the integrated high-level state to participate again in subsequent computation. Together, the two loops form a nested computational process consisting of fast local updates and slower global updates, allowing additional recurrent depth to operate on continuously evolving multimodal states.

\subsection{Why Need Loop?}
\label{sec:lv-2-3}

A natural question is what additional value recurrent computation provides compared with simply scaling up a conventional Transformer. To investigate this, we compare LoopVL with three non-recurrent Transformer-VL baselines under the same 0.14T-token training budget. All models adopt the same Transformer block design as LoopVL, differing primarily in the depth and width of the language backbone and in whether recurrent computation is used.

\textbf{Transformer-VL 1B} uses a standard Transformer backbone with 32 layers and a hidden size of 1536, matching LoopVL in both the number of unique Transformer layers and hidden dimension, while executing each layer only once during a forward pass. Transformer-VL 4B (Deep) scales the model primarily through depth, using 78 layers with a hidden size of 1792. Transformer-VL 4B (Wide) instead scales primarily through width, retaining 32 layers while increasing the hidden size to 2816.

\begin{table}[H]
\centering
\caption{Comparison of LoopVL and Transformer-VL baselines at 0.14T training tokens. Bold denotes the highest score in each benchmark.}
\label{tab:lv-recurrent-comparison}
\begingroup
\fontsize{8.25}{10.5}\selectfont
\setlength{\tabcolsep}{1.5pt}
\setlength{\extrarowheight}{0pt}
\setlength{\aboverulesep}{3pt}
\setlength{\belowrulesep}{3pt}
\renewcommand{\arraystretch}{1.4}
\newcommand{\lvrechead}[1]{{\sffamily\bfseries\fontsize{7.5}{9}\selectfont\makecell[c]{#1}}}
\newcommand{\lvrecbest}[1]{{\bfseries #1}}
\begin{tabular*}{\linewidth}{@{\extracolsep{\fill}}l*{11}{c}@{}}
\toprule
{\sffamily\bfseries\fontsize{7.5}{9}\selectfont Model} & \lvrechead{Recursions} &
\lvrechead{FLOPs\\($10^{21}$)} &
\lvrechead{Tokens\\(T)} &
\lvrechead{MMStar} & \lvrechead{RealWorld\\QA} &
\lvrechead{VMCBench} & \lvrechead{AI2D} &
\lvrechead{ChartQA} & \lvrechead{Math\\Vision} &
\lvrechead{VisuLogic} & \lvrechead{MMK12} \\
\midrule
{\bfseries LoopVL} & 4 (H2L3) & 2.47 & 0.14 &
\lvrecbest{63.47} & \lvrecbest{70.98} & 70.90 & 75.49 &
74.52 & \lvrecbest{38.49} & \lvrecbest{27.00} & \lvrecbest{49.65} \\
Transformer-VL 1B & 1 & 0.92 & 0.14 &
55.33 & 55.29 & 55.10 & 60.01 & 51.12 & 30.59 & 21.20 & 40.30 \\
Transformer-VL 4B (Deep) & 1 & 2.89 & 0.14 &
61.33 & 66.54 & 71.20 & \lvrecbest{76.13} & 75.24 & 35.92 & 26.60 & 48.35 \\
Transformer-VL 4B (Wide) & 1 & 2.99 & 0.14 &
60.47 & 66.01 & \lvrecbest{72.00} & 75.65 & \lvrecbest{75.56} & 33.55 & 26.30 & 47.70 \\
\bottomrule
\end{tabular*}
\endgroup
\end{table}

As shown in Table~\ref{tab:lv-recurrent-comparison}, the 32-layer Transformer-VL 1B baseline consistently underperforms LoopVL across all evaluated benchmarks. For example, LoopVL improves MMStar from 55.33 to 63.47, RealWorldQA from 55.29 to 70.98, and ChartQA from 51.12 to 74.52. This comparison directly demonstrates the effect of recurrent computation: LoopVL and Transformer-VL 1B contain the same number of unique Transformer layers, but LoopVL increases the executed depth from 32 to 128 layer calls by repeatedly reusing shared modules.

More importantly, LoopVL approaches or even surpasses substantially larger dense Transformers. Despite maintaining an approximately 1B-scale language backbone, LoopVL outperforms larger Transformer baselines while remaining competitive across a broad range of multimodal tasks. Meanwhile, the estimated training FLOPs of the two larger models are $2.89 \times 10^{21}$ and $2.99 \times 10^{21}$, respectively, compared with $2.47 \times 10^{21}$ for LoopVL.

These results illustrate the value of recurrent computation from a model-scaling perspective. Rather than increasing model capacity by storing substantially more independent parameters, LoopVL increases executed depth through repeated reuse of a compact parameter set. Under the same training-data budget, this parameter-sharing strategy enables a roughly 1B-scale language backbone to achieve multimodal performance comparable to dense Transformers several times larger.

\endgroup

\subsection{Why Loop Visual Tokens?}
\label{sec:lv-2-2}

LoopVL allows visual tokens to evolve throughout recurrent computation. Encoder features initially describe individual image patches, but their corresponding hidden states continue to be updated after entering the recurrent language backbone.

\begin{figure}[!t]
\centering
\includegraphics[width=\linewidth]{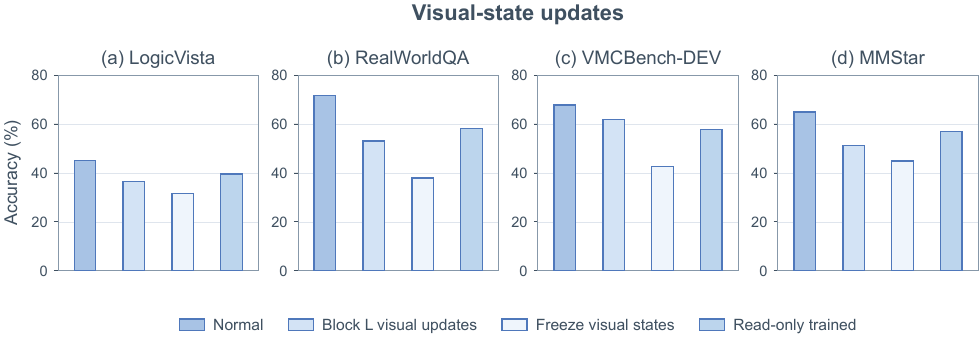}
\caption{\textbf{Visual-state update controls.} Normal is standard LoopVL training and inference. Block L visual updates and Freeze visual states are inference-only controls applied to the normal model: the former discards second-cycle L-module visual updates before they carry forward, whereas the latter keeps visual states fixed throughout the second cycle. Read-only trained is a separately trained variant that allows visual-state updates only in the first model cycle and uses the same read-only rule during training and inference. Visual tokens remain readable and nonvisual states continue updating.}
\label{fig:lv-visual-state-updates}
\end{figure}

Continued visual-state refinement. Recurrent computation allows the model to progressively select and organize visual evidence, updating patch representations and the relative importance of image regions across invocations. These updates act on evolving visual states without another pass through the visual encoder.

Visual revisiting under shared parameters. Parameter sharing fixes the transformation used at each recurrent step, while the state received by each invocation continues to change throughout the recurrent process. Once the first cycle has changed the visual state, the same shared module receives a different input in the next invocation and can therefore produce different visual updates and attention allocations. Because these visual positions maintain their spatial correspondence with image patches, we can also align the same physical layer across different cycles and directly observe how visual evidence is reallocated. This allows us to distinguish between ``simply repeating the same computation'' and ``applying shared parameters to continuously evolving visual states.''

Intervening on visual-state updates. Figure~\ref{fig:lv-visual-state-updates} compares four conditions on LogicVista, RealWorldQA, VMCBench-DEV, and MMStar. Normal uses standard LoopVL training and inference. Two inference-only controls leave the first cycle unchanged, then either discard L-module visual updates before they carry forward or keep visual states fixed during the second cycle. A fourth, separately trained variant allows visual-state updates only in the first cycle and applies the same read-only rule during training, validation, and generation. Normal achieves the highest accuracy on all four evaluations; the retrained read-only variant remains below it.

The retrained variant preserves the H2L3 schedule, both 16-layer shared stacks, and the final H-state readout. At the first-cycle boundary, it saves separate L/H visual-state references. During the second cycle, persistent visual states are restored to their respective references after writeback; within each module, visual residual states are restored after every attention and feed-forward update to a reference taken from that invocation's normally combined input. Visual tokens remain present and readable, while nonvisual states continue updating. The restriction adds no loss or extra visual-encoder pass and does not freeze second-cycle model parameters.


\section{Pre-Training}
\label{sec:lv-3}

\begin{figure}[H]
\centering
\includegraphics[width=\linewidth]{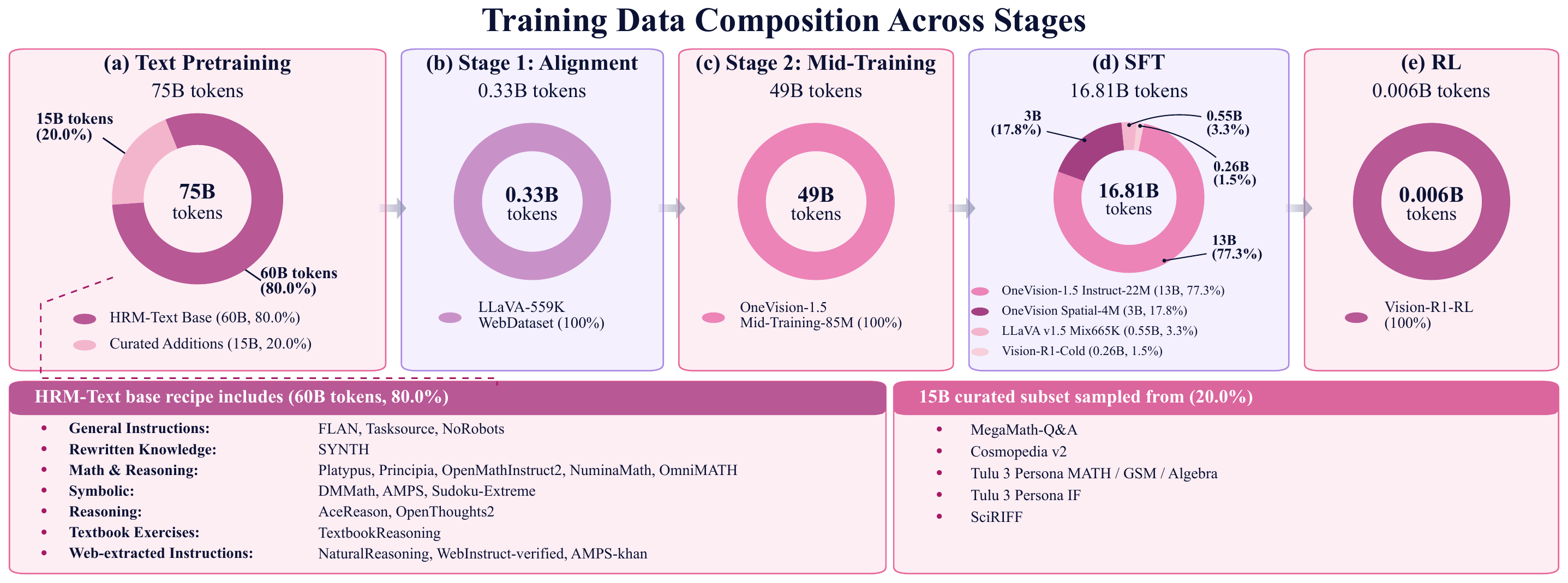}
\caption{Our training-data sources and token allocations across the LoopVL training stages. HRM-Text Base denotes the open-source training data recipe from HRM-Text used for our language pretraining.}
\label{fig:lv-training-data-composition}
\end{figure}

Figure~\ref{fig:lv-training-data-composition} summarizes our training data across the full pipeline, from language pretraining and multimodal adaptation to post-training. We pretrain our own language model based on HRM-Text's open-source framework, using its publicly released data recipe as the foundation and augmenting it with additional data curated and filtered by us. The text pretraining stage uses a total of 75B tokens, including 60B from the HRM-Text base data recipe and an additional 15B from the curated data shown in the figure.

Multimodal model construction then starts from our pretrained language-model checkpoint and the pretrained Penguin Vision Encoder. During Stage 1: Visual Language Alignment, we freeze both the language backbone and the Penguin Vision Encoder and train only the visual interface. During Stage 2: Multimodal Mid-Training, we jointly update the language model and visual interface while keeping the Penguin Vision Encoder frozen.

\subsection{Initialization}
\label{sec:lv-3-2}

The visual encoder retains its Penguin pretrained initialization, while the language backbone uses the H2L3 checkpoint obtained from our own text pretraining. In the dynamic-resolution Stage 1 configuration, the projector is randomly initialized and trained during Stage 1. The previous optimizer, scheduler, and trainer states are not loaded.

At the visual interface, we first apply parameter-free RMS normalization to the visual features, followed by calibration to the median RMS of our language embedding vectors. The visual representations then follow the standard backbone path with the inherited embedding scaling applied. Penguin and the language backbone use BF16, while the projector parameters remain in FP32. Before multimodal training, we perform regression tests to verify text-position compatibility, visual-slot counts, loss masks, and the intended frozen/trainable parameter boundaries.

\subsection{Learning Schedule}
\label{sec:lv-3-3}

Our main training recipes use AdamW with \((\beta_1,\beta_2)=(0.9,0.95)\), gradient-norm clipping at 1.0, and a learning-rate warmup covering 3\% of the total training steps. In Stage 1, the projector uses a reference learning rate of \(10^{-3}\). In Stage 2, the language backbone uses a learning rate of \(2\times10^{-5}\), while the projector uses \(10^{-4}\). Weight decay is set to 0 in the main training stages. Learning rates and frozen parameter groups are specified separately for each stage.

The global batch sizes are 256 and 128 for the two stages, respectively. Stage 1 reaches a global batch size of 256 through a combination of micro-batching and gradient accumulation. Stage 2 uses the corresponding micro-batch and gradient-accumulation configuration while maintaining a global batch size of 128. Gradient checkpointing is enabled during training, and inference caching is disabled. Stage 2 uses cosine learning-rate decay after warmup.

The active backward graph follows the selective recurrent schedule described in Section~\ref{sec:lv-2-1}. Multimodal training consistently follows this route. Length bucketing and batch-local padding reduce wasted sequence positions without changing the image-specific grid. During multimodal training, causal cross-entropy loss is computed only on valid response tokens, while prefix and padding positions are masked from supervision.

\subsection{Stage 1: Visual Language Alignment}
\label{sec:lv-3-4}

The alignment stage uses the LLaVA-559K WebDataset shown in Figure~\ref{fig:lv-training-data-composition}, with a token allocation of 0.33B. Under a maximum sequence length of 2048, each image uses 256--1024 visual tokens. The Penguin Vision Encoder and our language backbone remain frozen during this stage, while only the visual interface is optimized using image--caption supervision with loss applied to the response tokens. \citep{liu2023visual}

The goal of this stage is to establish a stable visual input pathway into the language model, allowing visual features to enter the frozen recurrent language backbone through the trainable interface. The retained input-gradient path enables the recurrent language backbone to propagate learning signals to the visual interface. Before the main alignment run, we perform a set of implementation checks to verify the intended frozen/trainable parameter boundaries and compare model responses under the original image input and image-replacement controls, confirming that the visual interface can learn and transmit image-dependent information.

\subsection{Stage 2: Multimodal Mid-Training}
\label{sec:lv-3-5}

Stage 2 starts from the aligned checkpoint without carrying over the optimizer or scheduler from the previous stage. The Penguin Vision Encoder remains frozen, while the language backbone and visual interface are trainable. The maximum sequence length is set to 3072, the global batch size is 128, and the visual-token range remains 256--1024. This stage is designed to adapt the recurrent model to instruction-conditioned multimodal answering across general visual questions, text-rich images, diagrams, mathematics, and short-answer tasks.

General visual-instruction supervision. Stage 2 uses OneVision-1.5 Mid-Training-85M, corresponding to the full 49B-token allocation shown in Figure~\ref{fig:lv-training-data-composition}. The language backbone and visual interface are jointly trained, enabling instruction-conditioned generation on top of the aligned visual input pathway while retaining the response-only training objective.

Answer preservation and format normalization. During data conversion, the original ground-truth responses are preserved. Instruction rewriting is used to normalize task phrasing, while OCR and document-related questions retain their original wording. For multiple-choice tasks, responses are standardized to an option letter together with the corresponding answer text; yes/no tasks use the appropriate short response. Image hashes and normalized question text are used to support deduplication.

\textbf{Long-response filtering.} A small fraction of text-rich samples contain responses that exceed the available sequence-length budget. During Stage 2 data preparation, these samples are removed from the training set rather than truncated or split into multiple segments. This filtering avoids introducing partial-response targets caused by length constraints.

Figure~\ref{fig:lv-training-data-composition} summarizes the data sources and token allocations across training stages, while Figure~\ref{fig:lv-training-workflow} presents the overall training workflow.


\section{Post-Training}
\label{sec:lv-4}

\begin{figure}[H]
\centering
\definecolor{lvFlowInk}{HTML}{29415B}
\definecolor{lvFlowBlue}{HTML}{5279A8}
\definecolor{lvFlowDeepBlue}{HTML}{36577F}
\definecolor{lvFlowTeal}{HTML}{356C64}
\definecolor{lvFlowNeutralPanel}{HTML}{F2F5FA}
\definecolor{lvFlowBluePanel}{HTML}{EAF1FA}
\definecolor{lvFlowTealPanel}{HTML}{EDF6F3}
\definecolor{lvFlowCanvas}{HTML}{F5F5F5}
\definecolor{lvFlowCanvasBorder}{HTML}{D9DDE2}
\resizebox{\linewidth}{!}{%
\begin{tikzpicture}[
  x=1cm, y=1cm,
  flow frame/.style={
    rectangle, rounded corners=3pt, line width=0.75pt,
    minimum width=2.25cm, minimum height=1.65cm,
    inner sep=0pt, outer sep=0pt
  },
  flow stage/.style={flow frame, draw=lvFlowBlue!60, fill=white, solid},
  flow model/.style={flow frame, draw=lvFlowDeepBlue, fill=lvFlowBluePanel,
    dash pattern=on 3pt off 2pt},
  flow label/.style={
    text=lvFlowInk, align=center, text width=2.05cm,
    inner sep=0pt, outer sep=0pt,
    font=\normalfont\fontsize{9.2}{11.3}\selectfont
  },
  flow model label/.style={flow label, text=lvFlowDeepBlue,
    font=\normalfont\bfseries\fontsize{10}{12}\selectfont},
  flow arrow/.style={-{Stealth[length=2mm,width=1.5mm]}, draw=lvFlowDeepBlue,
    line width=0.8pt, solid},
  flow heading/.style={font=\normalfont\sffamily\bfseries\fontsize{7.2}{8.5}\selectfont,
    text=lvFlowDeepBlue, inner sep=0pt, outer sep=0pt}
]
  \path[fill=lvFlowCanvas, draw=lvFlowCanvasBorder, line width=0.55pt,
    rounded corners=12pt] (-0.35,-0.48) rectangle (16.90,4.98);
  \path[fill=lvFlowNeutralPanel, rounded corners=7pt] (0,2.36) rectangle (5.70,4.50);
  \path[fill=lvFlowBluePanel, rounded corners=7pt] (2.92,0) rectangle (10.96,2.30);
  \path[fill=lvFlowTealPanel, rounded corners=7pt] (8.20,2.36) rectangle (16.55,4.50);
  \node[flow heading] at (2.85,4.33) {LANGUAGE BACKBONE};
  \node[flow heading] at (6.94,2.16) {MULTIMODAL ADAPTATION};
  \node[flow heading,text=lvFlowTeal] at (12.375,4.33) {POST-TRAINING};

  \node[flow stage] (pretrain) at (1.62,3.325) {};
  \node[flow model] (llm) at (4.28,3.325) {};
  \node[flow stage] (align) at (4.28,1.175) {};
  \node[flow stage] (midtrain) at (6.94,1.175) {};
  \node[flow model] (base) at (9.60,1.175) {};
  \node[flow stage,draw=lvFlowTeal!65] (sft) at (9.60,3.325) {};
  \node[flow stage,draw=lvFlowTeal!75] (rl) at (12.26,3.325) {};
  \node[flow model,draw=lvFlowTeal,fill=lvFlowTealPanel] (loopvl) at (14.92,3.325) {};

  \node[flow label] at (pretrain.center) {Pre-Training};
  \node[flow model label] at (llm.center) {LLM};
  \node[flow label] at (align.center) {Stage 1:\\Visual\\Language\\Alignment};
  \node[flow label] at (midtrain.center) {Stage 2:\\Multimodal\\Mid-Training};
  \node[flow model label] at (base.center) {LoopVL\\Base};
  \node[flow label] at (sft.center) {Supervised\\Fine-Tuning};
  \node[flow label] at (rl.center) {Visual\\Reinforcement\\Learning};
  \node[flow model label,text=lvFlowTeal] at (loopvl.center) {LoopVL};

  \draw[flow arrow] (pretrain.east) -- (llm.west);
  \draw[flow arrow] (llm.south) -- (align.north);
  \draw[flow arrow] (align.east) -- (midtrain.west);
  \draw[flow arrow] (midtrain.east) -- (base.west);
  \draw[flow arrow,draw=lvFlowTeal] (base.north) -- (sft.south);
  \draw[flow arrow,draw=lvFlowTeal] (sft.east) -- (rl.west);
  \draw[flow arrow,draw=lvFlowTeal] (rl.east) -- (loopvl.west);
\end{tikzpicture}%
}
\caption{Overview of our LoopVL training workflow. Dashed boxes denote models; solid boxes denote training steps. Language pretraining produces our own LLM checkpoint. Visual--language alignment and multimodal mid-training form LoopVL-Base, followed by supervised fine-tuning and visual reinforcement learning. Data sources and token allocations are shown in Figure~\ref{fig:lv-training-data-composition}.}
\label{fig:lv-training-workflow}
\end{figure}
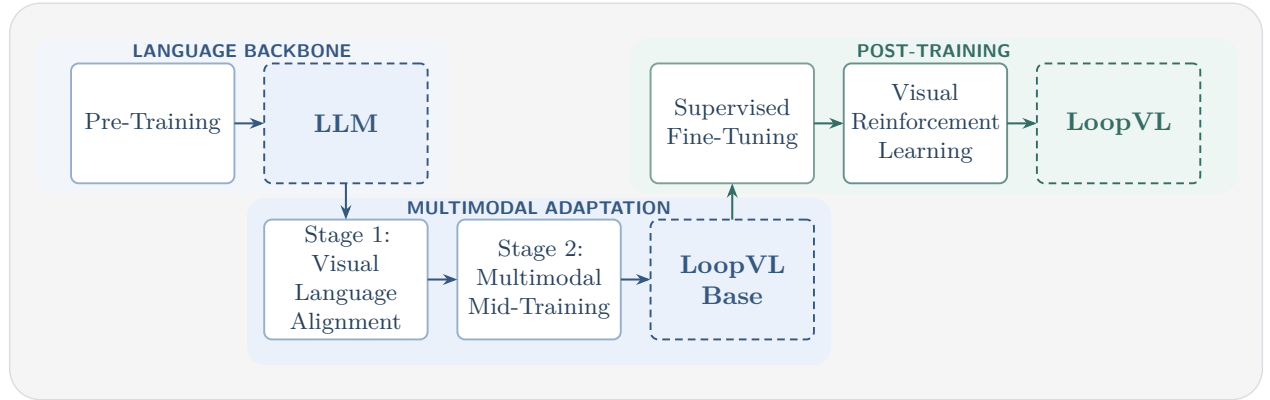

After multimodal mid-training, we continue with supervised fine-tuning and visual reinforcement learning. Figure~\ref{fig:lv-training-workflow} illustrates this training pipeline. Supervised fine-tuning uses 16.81B tokens from four data sources, followed by a reinforcement-learning stage using 0.006B tokens from Vision-R1-RL. These stages continue from the model built with our own pretrained language backbone and visual interface.

\subsection{Supervised Fine-Tuning}
\label{sec:lv-4-1}

\textbf{Data mixture.} The supervised fine-tuning stage uses a total of 16.81B tokens, combining OneVision-1.5 Instruct-22M (13B tokens), OneVision Spatial-4M (3B), LLaVA v1.5 Mix665K (0.55B), and Vision-R1-Cold (0.26B). As shown in Figure~\ref{fig:lv-training-data-composition}, they account for approximately 77.3\%, 17.8\%, 3.3\%, and 1.5\%, respectively. \citep{liu2024improved}

\textbf{Supervised objective.} Training continues to use a PrefixLM mask. Image and instruction positions jointly form a bidirectional conditioning prefix, while response positions can attend to this prefix and to preceding response tokens. Let \(x\) denote the conditioning context containing the image and instruction, and let \(y=(y_1,\ldots,y_T)\) denote the reference response. The conditional generation probability is factorized as

\begin{equation}
p_{\theta}(y\mid x)
= \prod_{t=1}^{T} p_{\theta}\!\left(y_t\mid x,y_{<t}\right).
\label{eq:lv-autoregressive-probability}
\end{equation}

We train with teacher forcing, minimizing mean cross-entropy over valid response tokens across a batch of \(B\) samples and masking padding positions:

\begin{equation}
\mathcal{L}_{\mathrm{SFT}}(\theta)
= -\frac{
\displaystyle\sum_{i=1}^{B}\sum_{t=1}^{T_i}
\log p_{\theta}\!\left(y_{i,t}\mid x_i,y_{i,<t}\right)
}{
\displaystyle\sum_{i=1}^{B}T_i
}.
\label{eq:lv-sft-objective}
\end{equation}

Here, \(T_i\) denotes the number of valid response tokens in the \(i\)-th sample, and \(\theta\) denotes the trainable model parameters.

\subsection{Visual Reinforcement Learning}
\label{sec:lv-4-2}

This stage applies Group Relative Policy Optimization (GRPO) for visual reinforcement learning, combining an answer-correctness reward with a progressive reasoning-length schedule to further improve the model's multimodal reasoning capability.

\textbf{Outcome reward and group-relative policy optimization.} For each image--question input, the model samples a group of responses, assigning positive rewards when final answers match the reference answer and zero otherwise. GRPO computes advantages from group-normalized rewards and updates the policy using probability-ratio clipping and KL regularization against a reference policy.

\textbf{Progressive reasoning-length schedule.} We adopt Progressive Thinking Suppression Training, increasing the maximum generation length from 2048 to 4096 tokens. This schedule is designed to encourage compact correct reasoning initially and provide more room for complex derivations later. As the generation limit increases, we reduce the number of candidates per input to balance response length and candidate count.


\section{Evaluation}
\label{sec:lv-5}

\begingroup
\clubpenalties 2 10000 0
\widowpenalties 2 10000 0

Section~\ref{sec:lv-5-1} reports overall benchmark results. Sections~\ref{sec:lv-5-2} and~\ref{sec:lv-5-3} examine training-time recurrence configurations and inference-time schedules, respectively. Section~\ref{sec:lv-5-6} analyzes Visual Aha Moments, Section~\ref{sec:lv-5-5} examines cross-loop token relevance, and Section~\ref{sec:lv-5-4} studies visual-attention allocation. Finally, Section~\ref{sec:lv-5-7} investigates visual-state refinement and prediction formation.

\subsection{Results}
\label{sec:lv-5-1}

\textbf{Evaluation protocol.} We evaluate LoopVL across 16 benchmark families spanning general understanding, hallucination, mathematics, and science. Our main evaluation uses greedy generation and, depending on the task, asks the model to produce an option letter, a yes/no answer, or a short final response without requiring an explicit reasoning trace. For short-answer tasks, we typically set the maximum generation length to 32 tokens. When chain-of-thought reasoning is explicitly elicited through prompting, we observe comparable or slightly higher performance on mathematics and reasoning tasks; in contrast, explicit chain-of-thought tends to reduce accuracy on short-answer tasks that can be answered directly.

\begin{table}[!t]
\centering
\caption{Multimodal comparison of LoopVL-1B and compact vision--language models. The display contains 16 benchmark families and 23 score rows when POPE categories and MMK12 subjects are expanded. Darker and lighter blue indicate the best and second-best scores, respectively. Entries combine in-house evaluations with selected public results.}
\label{tab:model-comparison}
\label{tab:loopvl-multimodal}
\label{tab:lv-main}
\begingroup
\colorlet{loopvlbestbg}{ai2pink!30}
\colorlet{loopvlsecondbg}{ai2pink!12}
\newcommand{\loopvlbest}[1]{\cellcolor{loopvlbestbg}{\bfseries #1}}
\newcommand{\loopvlsecond}[1]{\cellcolor{loopvlsecondbg}#1}
\fontsize{8.2}{10.6}\selectfont
\setlength{\tabcolsep}{2.6pt}
\setlength{\extrarowheight}{0pt}
\renewcommand{\arraystretch}{1}
\newlength{\loopvlmodelwidth}
\setlength{\loopvlmodelwidth}{\dimexpr(1.1\linewidth-26.5mm-25\tabcolsep-\arrayrulewidth)/12\relax}
\newcommand{\loopvlrowstrut}{\rule[-4.5pt]{0pt}{15pt}}
\newcommand{\loopvlhead}[1]{\parbox[c][36pt][c]{\loopvlmodelwidth}{\centering\sffamily\bfseries\color{black}\fontsize{6.6}{8.2}\selectfont#1\par}}
\newcommand{\loopvlwildlabel}{\raisebox{1pt}[0pt][0pt]{\parbox[c]{26.5mm}{\raggedright\fontsize{7.4}{7.4}\selectfont MathVision-\\WildPhoto}}}
\makebox[\linewidth][c]{%
\begin{tabular}{@{}>{\loopvlrowstrut\raggedright\arraybackslash}p{26.5mm}|*{12}{>{\centering\arraybackslash}p{\loopvlmodelwidth}}}
\hline
\multicolumn{1}{@{}c|}{\parbox[c][36pt][c]{26.5mm}{\centering\sffamily\bfseries\color{black}\fontsize{6.6}{8.2}\selectfont Benchmark}} & \loopvlhead{LoopVL\\1B} &
\loopvlhead{Qwen3.5\\2B} &
\loopvlhead{LFM2-VL\\3B} &
\loopvlhead{gemma-4\\E2B-it} &
\loopvlhead{gemma-4\\E4B-it} &
\loopvlhead{InternVL\\3.5-2B} &
\loopvlhead{InternVL\\3.5-4B} &
\loopvlhead{LFM2.5\\VL-1.6B} &
\loopvlhead{PLaMo2.1\\2B-VL} &
\loopvlhead{MiniCPM\\V-4.6} &
\loopvlhead{Ministral\\3-3B\\Instruct} &
\loopvlhead{jina-\\vlm} \\
\hline
Training tokens & \LoopVLTrainingTokens & 36T & 11.1T & 11T & 11T & 36.38T & 36.38T & 28.1T & 3T & 36.1T & 13T & 36T \\
\hline
{\sffamily\bfseries General} &  &  &  &  &  &  &  &  &  &  &  &  \\
BabyVision & 13.14 & 11.60 & \loopvlbest{20.10} & 12.37 & \loopvlsecond{17.53} & 13.40 & 16.49 & 11.08 & 15.72 & 11.60 & 12.89 & 14.43 \\
LogicVista & \loopvlbest{45.19} & 30.65 & 34.23 & 27.29 & 33.78 & 31.32 & \loopvlsecond{36.91} & 29.31 & 25.06 & 30.87 & 32.66 & 31.32 \\
MMEval-Pro & 24.62 & 24.43 & \loopvlsecond{35.93} & 22.21 & 30.05 & 32.73 & \loopvlbest{42.82} & 26.32 & 11.02 & 31.78 & 30.25 & 33.82 \\
MMMU-Pro & \loopvlbest{38.67} & 24.90 & 27.98 & 26.76 & 33.06 & 31.33 & \loopvlsecond{38.30} & 24.22 & 13.06 & 26.24 & 29.77 & 26.71 \\
MMStar & \loopvlbest{63.47} & 55.10 & 57.00 & 42.13 & 51.13 & 58.53 & \loopvlsecond{63.20} & 48.53 & 33.27 & 57.13 & 48.33 & 56.40 \\
RealWorldQA & \loopvlbest{70.98} & \loopvlsecond{69.00} & 68.63 & 45.10 & 55.29 & 59.87 & 65.10 & 63.40 & 46.93 & 64.71 & 56.47 & 60.52 \\
VMCBench & 70.90 & 75.10 & \loopvlsecond{76.20} & 65.50 & 72.70 & 73.60 & \loopvlbest{78.10} & 69.30 & 49.40 & 75.60 & 68.50 & 73.90 \\
VisuLogic & \loopvlbest{27.00} & \loopvlsecond{26.80} & 26.20 & 25.00 & 23.20 & 22.10 & 25.70 & 24.70 & 21.90 & 25.60 & 26.50 & 26.50 \\
VisualPuzzles & \loopvlsecond{34.76} & \loopvlbest{38.30} & 28.94 & 27.14 & 31.85 & 28.17 & 30.31 & 25.17 & 22.60 & 26.63 & 29.02 & 26.88 \\
\hline
{\sffamily\bfseries Hallucination} &  &  &  &  &  &  &  &  &  &  &  &  \\
HallusionBench & 44.71 & 46.95 & 45.97 & 39.81 & 48.96 & 47.02 & \loopvlbest{51.59} & 40.27 & 32.99 & \loopvlsecond{49.71} & 45.74 & 36.63 \\
POPE-A & 87.60 & 86.57 & \loopvlsecond{87.62} & 74.31 & 84.24 & 85.75 & 85.95 & 86.88 & 79.61 & 87.60 & 83.28 & \loopvlbest{88.31} \\
POPE-P & \loopvlbest{91.92} & 88.56 & 88.82 & 75.16 & 86.56 & 88.17 & 88.67 & 88.40 & 80.42 & 88.52 & 84.59 & \loopvlsecond{89.64} \\
POPE-R & \loopvlsecond{92.78} & 92.02 & 90.93 & 75.63 & 88.73 & \loopvlbest{93.41} & 92.19 & 89.59 & 81.15 & 90.59 & 87.40 & 91.33 \\
POPE & \loopvlbest{90.71} & 88.60 & 89.10 & 75.03 & 86.47 & 89.00 & 88.86 & 88.28 & 80.39 & 88.89 & 85.05 & \loopvlsecond{89.74} \\
\hline
{\sffamily\bfseries Mathematics} &  &  &  &  &  &  &  &  &  &  &  &  \\
ChartQA & 74.52 & 77.50 & \loopvlsecond{82.32} & 47.24 & 44.88 & 79.64 & \loopvlbest{86.00} & 73.16 & 30.60 & 67.36 & 78.96 & 81.40 \\
MMK12-Math & \loopvlbest{61.60} & 35.00 & 38.80 & 32.60 & 37.40 & 42.40 & \loopvlsecond{54.80} & 30.80 & 26.00 & 29.20 & 33.00 & 35.80 \\
MathVision & \loopvlbest{38.49} & \loopvlsecond{38.40} & 18.72 & 16.35 & 19.64 & 17.30 & 22.04 & 16.51 & 9.05 & 8.72 & 18.32 & 19.67 \\
\loopvlwildlabel & 20.39 & 22.04 & 21.05 & \loopvlsecond{22.70} & 19.41 & \loopvlbest{27.30} & \loopvlsecond{22.70} & 13.49 & 15.79 & 19.41 & 21.05 & 20.39 \\
\hline
{\sffamily\bfseries Science} &  &  &  &  &  &  &  &  &  &  &  &  \\
AI2D & 75.49 & 75.20 & 79.18 & 65.54 & 75.55 & 78.72 & \loopvlbest{82.80} & 71.44 & 41.97 & 80.47 & 77.36 & \loopvlsecond{82.06} \\
MMK12-Biology & 46.20 & 29.00 & 35.00 & 39.20 & 39.40 & \loopvlsecond{53.80} & \loopvlbest{58.00} & 31.40 & 29.40 & 39.80 & 35.00 & 43.80 \\
MMK12-Chemistry & 48.00 & 24.20 & 32.60 & 33.00 & 32.80 & \loopvlsecond{53.60} & \loopvlbest{66.60} & 23.00 & 29.80 & 37.80 & 33.60 & 39.60 \\
MMK12-Physics & \loopvlsecond{42.80} & 29.00 & 34.00 & 33.20 & 33.20 & 42.60 & \loopvlbest{45.00} & 29.20 & 31.80 & 32.60 & 32.20 & 33.80 \\
MMK12 & \loopvlsecond{49.65} & 29.30 & 35.10 & 34.50 & 35.70 & 48.10 & \loopvlbest{56.10} & 28.60 & 29.25 & 34.85 & 33.45 & 38.25 \\
\hline
\end{tabular}
}%
\par
\endgroup
\end{table}

We use VLMEvalKit and official benchmark code and evaluation scripts; Appendix~\ref{app:lv-evaluation} specifies the tools used for each benchmark. Metrics follow the official definitions of each benchmark. ChartQA uses relaxed accuracy; HallusionBench reports the average of aAcc, fAcc, and qAcc; MMEval-Pro uses a source-level macro-average of genuine accuracy; POPE is reported as pooled F1 over category instances; and MMK12 reports the accuracies of four 500-question subjects together with the overall accuracy.

Table~\ref{tab:lv-main} compares LoopVL-1B with vision--language models of broadly comparable parameter scale, showing strong performance on visual reasoning and general visual understanding while remaining competitive on hallucination, chart understanding, and scientific reasoning. These results motivate our analysis of how recurrent scheduling and evolving visual states contribute to multimodal computation.

\subsection{Ablations}
\label{sec:lv-5-2}

\textbf{Training with different recurrence configurations.} To investigate how the organization of recurrent computation affects model performance, we independently train H1L1, H1L3, H2L1, and H2L3 configurations from initialization. Each model uses its designated H/L recurrence structure from the language pre-training stage onward and maintains the same computation pattern throughout visual--language alignment, multimodal training, and post-training.

Table~\ref{tab:lv-training-ablation} shows that performance depends strongly on the H/L recurrence organization. H2L3 achieves the highest score on all five benchmarks. Notably, H1L3 and H2L1 both execute 64 Transformer-layer calls, yet H2L1 performs better on all five tasks. This indicates that even at the same unrolled computational depth, different L/H update orders can lead to different performance. Model quality therefore depends not only on how much computation is executed, but also on how that computation is organized between low-level state refinement and high-level state updates.

\begin{table}[H]
\centering
\begin{minipage}{0.94\linewidth}
\caption{Training-time H/L configuration ablation. Each variant is independently trained from initialization using its designated recurrence schedule throughout the full training pipeline. Unrolled layers denote the number of Transformer-layer calls in one forward pass.}
\label{tab:loopvl-training-ablation}
\label{tab:lv-training-ablation}
\begingroup
\fontsize{9}{11}\selectfont
\setlength{\tabcolsep}{3.8pt}
\setlength{\extrarowheight}{0pt}
\renewcommand{\arraystretch}{1}
\newlength{\loopvltrainingwidth}
\setlength{\loopvltrainingwidth}{\dimexpr(\linewidth-28mm-12mm-14\tabcolsep)/5\relax}
\newcommand{\loopvltrainingstrut}{\rule[-5.5pt]{0pt}{18pt}}
\newcommand{\loopvltraininghead}[2]{\parbox[c][27pt][c]{#1}{\centering\sffamily\bfseries\color{black}\fontsize{8.2}{10}\selectfont#2\par}}
\begin{tabular}{>{\loopvltrainingstrut\raggedright\arraybackslash}p{28mm}>{\centering\arraybackslash}p{12mm}*{5}{>{\centering\arraybackslash}p{\loopvltrainingwidth}}}
\hline
\loopvltraininghead{28mm}{Configuration} &
\loopvltraininghead{12mm}{Unrolled\\layers} &
\loopvltraininghead{\loopvltrainingwidth}{MMStar} &
\loopvltraininghead{\loopvltrainingwidth}{RealWorldQA} &
\loopvltraininghead{\loopvltrainingwidth}{VMCBench} &
\loopvltraininghead{\loopvltrainingwidth}{AI2D} &
\loopvltraininghead{\loopvltrainingwidth}{ChartQA} \\
\hline
H1L1 & 32 & 55.33 & 55.29 & 55.10 & 60.01 & 51.12 \\
H1L3 & 64 & 58.13 & 59.61 & 60.50 & 66.06 & 60.40 \\
H2L1 & 64 & 60.80 & 64.58 & 63.80 & 68.26 & 65.12 \\
H2L3 & 128 & \textbf{63.47} & \textbf{70.98} & \textbf{70.90} & \textbf{75.49} & \textbf{74.52} \\
\hline
\end{tabular}
\par
\endgroup
\end{minipage}
\end{table}

\subsection{Loop Depth}
\label{sec:lv-5-3}

\textbf{Inference-time schedule sensitivity.} Table~\ref{tab:lv-inference-ablation} fixes the trained H2L3 model and changes only the H/L configuration used at inference. Unlike the independently trained variants in Section~\ref{sec:lv-5-2}, these configurations reuse the same parameters without retraining. The training-matched H2L3 schedule achieves the highest score on all five benchmarks. Under H1L1--H1L3, reading out from earlier hidden states yields largely unparseable responses and near-zero scores. These results characterize early readout from the H2L3-trained model, rather than the capabilities of models independently trained with a single model cycle.

With the number of model cycles fixed at two, increasing the L invocations per cycle from one to three improves performance on all five tasks. The comparison changes the amount of low-level refinement between H updates without adding parameters or changing the number of model cycles. For example, MMStar rises from 29.40 under H2L1 to 51.33 under H2L2 and 63.47 under H2L3. Thus, reducing low-level update calls at inference does not preserve this checkpoint's performance, even when the number of H updates remains unchanged.

Increasing the number of model cycles beyond the training configuration does not produce a corresponding gain. Holding L3 fixed, H3L3 and H4L3 execute 192 and 256 Transformer-layer calls, respectively, compared with 128 under H2L3. Nevertheless, MMStar declines from 63.47 to 58.20 and 24.00, with the same ordering across all five benchmarks. Additional model cycles therefore do not automatically improve the results of this checkpoint, despite increasing the amount of executed computation.

\textbf{Equal unrolled depth does not imply equivalent behavior.} H1L3 and H2L1 both execute 64 Transformer-layer calls, H2L2 and H3L1 both execute 96, and H2L3 and H4L1 both execute 128, yet their performance differs substantially. For the 128-layer pair, MMStar is 63.47 under H2L3 but only 24.67 under H4L1. Within each pair, the schedules change the allocation and order of L/H updates while preserving parameters and executed depth. Models trained with longer schedules may behave differently. Taken together, the tested schedules show that this model performs best when training and inference recurrence configurations are matched; increasing depth or redistributing calls alone does not recover the training-matched performance.

\begin{table}[H]
\centering
\begin{minipage}{0.94\linewidth}
\caption{Inference-time H/L schedule sweep for the H2L3-trained model. H denotes the number of complete model cycles, and L denotes the number of L-module invocations within each cycle. Unrolled depth counts the total number of executed Transformer-layer calls in one forward pass. Darker and lighter shading indicate the best and second-best results, respectively.}
\label{tab:loopvl-hl-ablation}
\label{tab:lv-inference-ablation}
\begingroup
\definecolor{loopvlablationgreen}{HTML}{E8F1EC}
\colorlet{loopvlablationsecondgreen}{loopvlablationgreen!55}
\newcommand{\loopvlablationsecond}[1]{\cellcolor{loopvlablationsecondgreen}#1}
\fontsize{9}{11}\selectfont
\setlength{\tabcolsep}{3.8pt}
\setlength{\extrarowheight}{0pt}
\renewcommand{\arraystretch}{1}
\newlength{\loopvlablationwidth}
\setlength{\loopvlablationwidth}{\dimexpr(\linewidth-28mm-12mm-14\tabcolsep)/5\relax}
\newcommand{\loopvlablationstrut}{\rule[-4.5pt]{0pt}{16pt}}
\newcommand{\loopvlablationhead}[2]{\parbox[c][25pt][c]{#1}{\centering\sffamily\bfseries\color{black}\fontsize{8.2}{10}\selectfont#2\par}}
\begin{tabular}{>{\loopvlablationstrut\raggedright\arraybackslash}p{28mm}>{\centering\arraybackslash}p{12mm}*{5}{>{\centering\arraybackslash}p{\loopvlablationwidth}}}
\hline
\loopvlablationhead{28mm}{Configuration} &
\loopvlablationhead{12mm}{Unrolled\\layers} &
\loopvlablationhead{\loopvlablationwidth}{MMStar} &
\loopvlablationhead{\loopvlablationwidth}{RealWorldQA} &
\loopvlablationhead{\loopvlablationwidth}{VMCBench} &
\loopvlablationhead{\loopvlablationwidth}{AI2D} &
\loopvlablationhead{\loopvlablationwidth}{ChartQA} \\
\hline
H1L1 & 32 & 0.47 & 0.00 & 0.50 & 1.62 & 0.00 \\
H1L2 & 48 & 0.07 & 0.00 & 0.30 & 0.06 & 0.00 \\
H1L3 & 64 & 0.07 & 0.00 & 0.10 & 0.45 & 0.04 \\
\hline
H2L1 & 64 & 29.40 & 12.81 & 27.90 & 25.74 & 17.40 \\
H2L2 & 96 & 51.33 & 58.95 & 63.10 & 65.38 & \loopvlablationsecond{63.28} \\
\rowcolor{loopvlablationgreen}
\textbf{H2L3} & 128 & \textbf{63.47} & \textbf{70.98} & \textbf{70.90} & \textbf{75.49} & \textbf{74.52} \\
\hline
H3L1 & 96 & 34.27 & 29.67 & 30.00 & 32.90 & 16.44 \\
H3L2 & 144 & 51.33 & 46.41 & 59.70 & 63.24 & 50.64 \\
H3L3 & 192 & \loopvlablationsecond{58.20} & \loopvlablationsecond{60.13} & \loopvlablationsecond{65.70} & \loopvlablationsecond{67.78} & 56.80 \\
\hline
H4L1 & 128 & 24.67 & 13.07 & 25.50 & 20.17 & 1.52 \\
H4L2 & 192 & 32.13 & 9.28 & 32.60 & 32.84 & 14.68 \\
H4L3 & 256 & 24.00 & 14.51 & 22.50 & 21.66 & 12.96 \\
\hline
\end{tabular}
\par
\endgroup
\end{minipage}
\end{table}

\subsection{Visual Aha Moment}
\label{sec:lv-5-6}

For one query and attention head, let \(a_i\) denote the attention weight assigned to visual token \(i\). All sums below are taken over the \(N\) visual tokens, and we define
\begin{equation}
p_i=\frac{a_i}{\sum_j a_j},
\label{eq:lv-visual-attention-normalization}
\end{equation}
which gives the attention distribution normalized over the image.

Normalized spatial entropy measures how broadly visual attention is distributed:
\begin{equation}
E(p)=-\frac{1}{\log N}\sum_i p_i\log p_i.
\label{eq:lv-spatial-entropy}
\end{equation}
Lower entropy indicates greater concentration. We use natural logarithms and the convention \(0\log0=0\). Entropy is first computed for each reference-answer query and then averaged equally over queries, heads, and samples.

\begin{figure}[!htbp]
\centering
\includegraphics[width=0.94\linewidth]{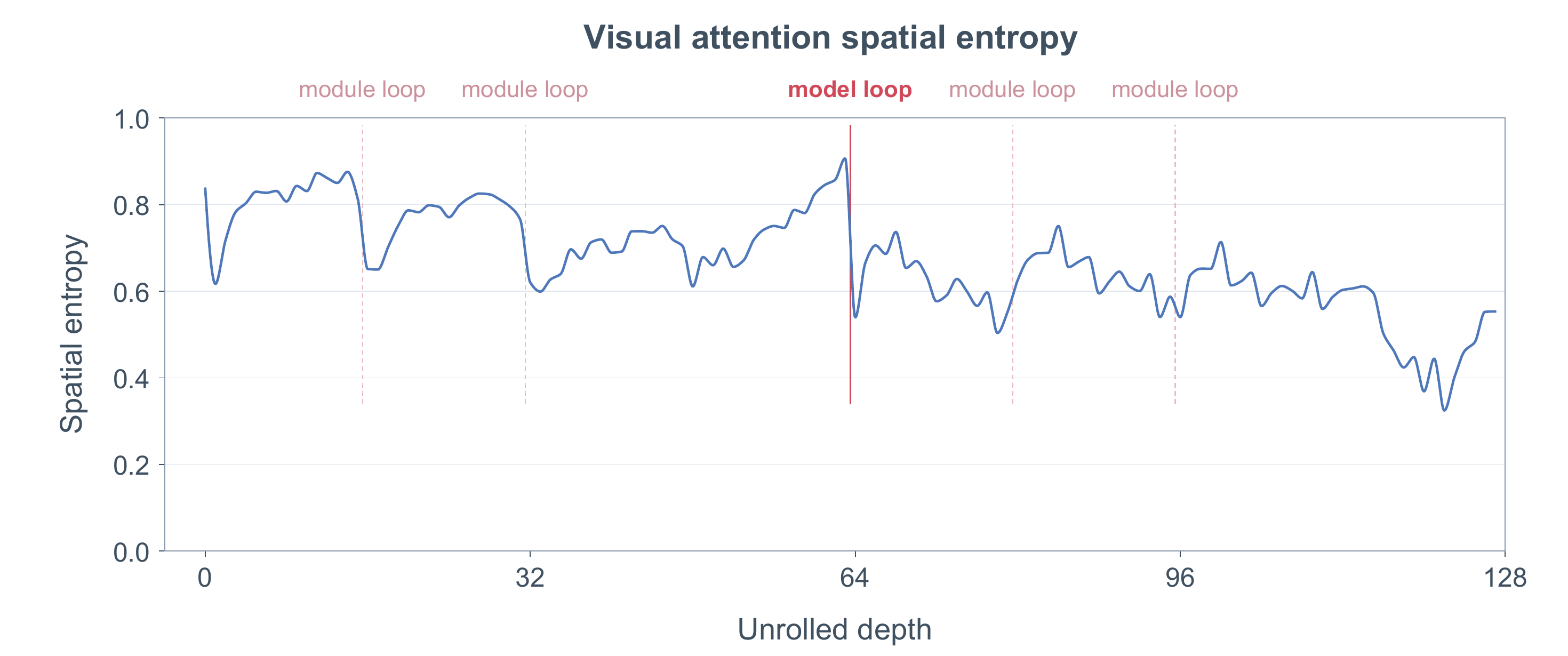}
\caption{Spatial entropy over recurrent depth. The curve reports mean answer-side visual-attention entropy for the 32-sample diagnostic set. Lower values indicate a more concentrated spatial distribution. Entropy is generally lower during the second cycle, but fluctuates across module calls rather than decreasing monotonically. The horizontal axis uses zero-based executed-layer indices; loop markers identify repeated L calls and the cycle boundary.}
\label{fig:lv-entropy-trajectory}
\end{figure}

\begin{figure}[!ht]
\centering
\includegraphics[width=0.94\linewidth]{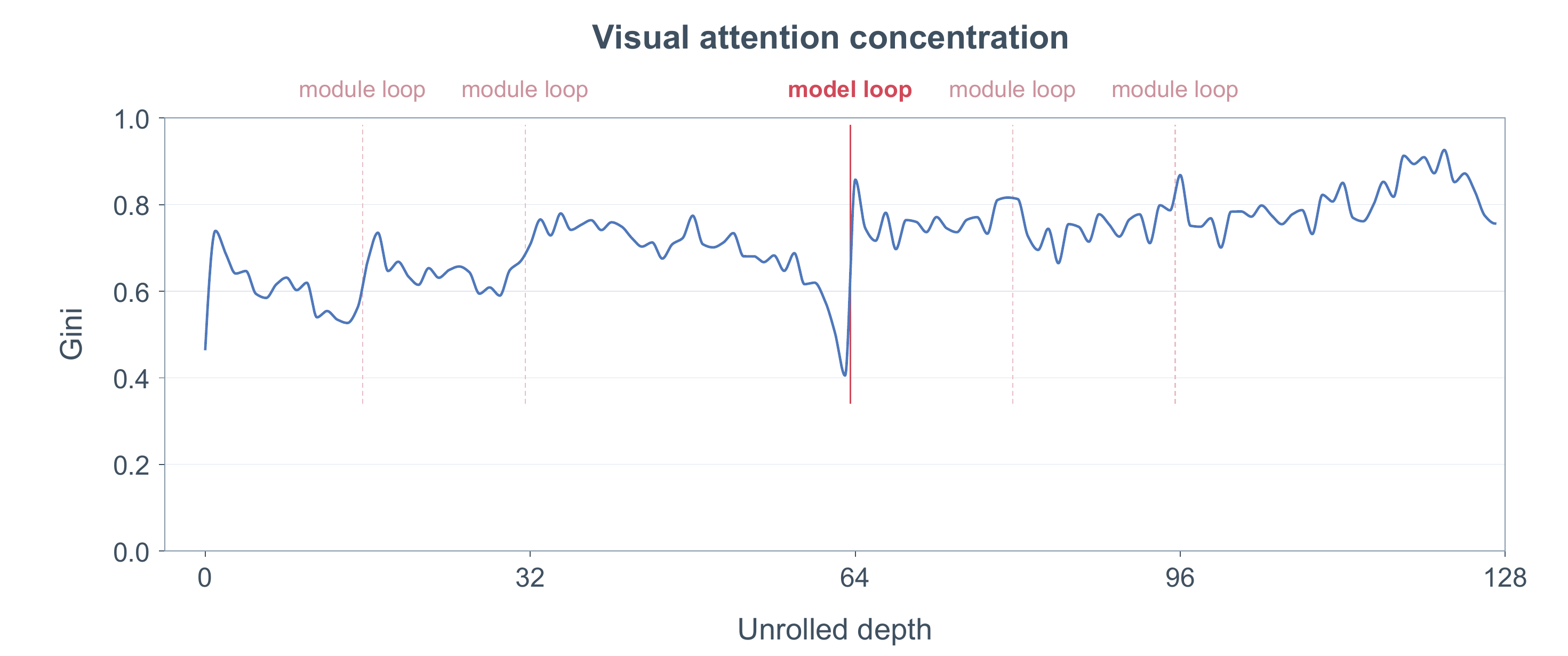}
\caption{Timing of visual-attention concentration across the recurrent trajectory. The mean Gini curve uses instruction and reference-answer query positions in the 32-sample controlled diagnostic set. The marked model-cycle transition lies between zero-based indices 63 and 64, where mean Gini rises from 0.405 to 0.858. A high Gini value describes concentration; it does not by itself identify attention sinks or causally important visual tokens.}
\label{fig:lv-gini-trajectory}
\end{figure}

Gini concentration measures the inequality among visual-token attention weights:
\begin{equation}
G(r)=\frac{1}{2N}\sum_{i,j}|r_i-r_j|.
\label{eq:lv-attention-gini}
\end{equation}
For each attention head, \(r\) is obtained by first averaging raw attention over instruction and reference-answer queries and then normalizing over visual tokens. Gini is then computed on this distribution and averaged over heads and samples. Higher Gini indicates more concentrated visual attention.

Figures~\ref{fig:lv-entropy-trajectory} and~\ref{fig:lv-gini-trajectory} apply the entropy and Gini definitions in Equations~\eqref{eq:lv-spatial-entropy} and~\eqref{eq:lv-attention-gini}, respectively, at every executed Transformer layer using their corresponding query scopes. The alignment between the model-cycle boundary and the change in attention concentration indicates that the shared modules exhibit substantially different visual reading patterns when entering the next recurrent cycle.

\textbf{Backpropagation-route robustness.} To rule out the possibility that this phenomenon is caused solely by the selective placement of backpropagation, we additionally pretrain a model from scratch with gradients propagated through all recurrent invocations. This model reproduces the cross-cycle attention shift and boundary concentration patterns observed in Figures~\ref{fig:lv-entropy-trajectory} and~\ref{fig:lv-gini-trajectory}, indicating that the phenomenon does not depend on the default selective backpropagation route.We further pretrain another model from scratch with the visual anchor and query-conditioned visual gate removed. The same cross-cycle attention shift and boundary concentration pattern remain, indicating that the phenomenon is not specific to the visual anchor–gate mechanism.

Taken together, these results show that the model-cycle transition is a stable point at which the organization of visual attention changes markedly. We refer to this pronounced cross-cycle shift in visual evidence allocation as a Visual Aha Moment.

\subsection{Cross-Loop Token Relevance}
\label{sec:lv-5-5}

Greater attention concentration alone is not sufficient to show that a later loop reads the same evidence more effectively. Another possibility is that the relative ranking of visual locations changes: patches that receive relatively little attention in the first cycle may receive substantially more attention in the second. Figure~\ref{fig:lv-reallocation-cases} examines this possibility in eight selected cases by aligning the same H layer at the endpoints of the two cycles.

\begin{figure}[!htbp]
\centering
\includegraphics[width=\linewidth]{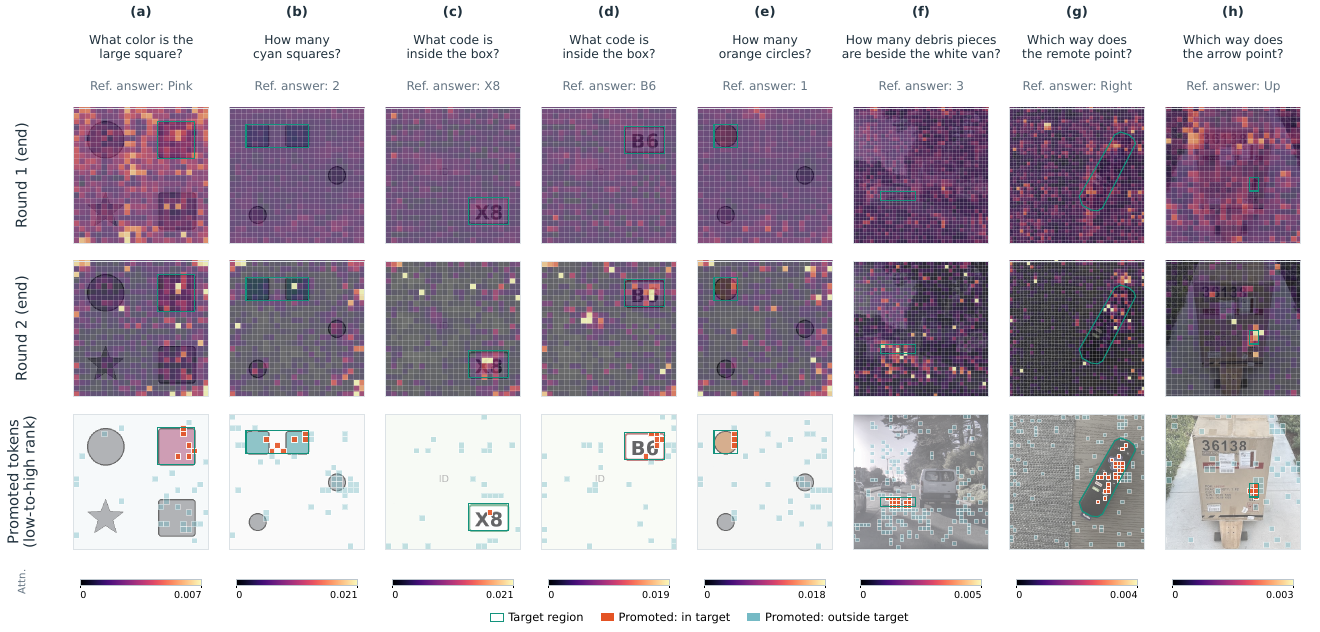}
\caption{Cross-cycle visual reallocation in eight selected examples. The first two rows compare the same final H-module layer at the two cycle endpoints (zero-based indices 63 and 127). The third marks tokens promoted from the first-cycle bottom half to the second-cycle top quartile: orange inside the outlined target region and cyan outside it. Each case uses a shared attention color scale across its two rounds. Reference answers identify the task, not intermediate model predictions.}
\label{fig:lv-reallocation-cases}
\end{figure}

\textbf{Rank reallocation in controlled visual examples.} In the color-recognition and cyan-square counting examples, the attention assigned to the annotated target region increases across cycles. Some positions in the second-cycle top quartile belonged to the bottom half of the first-cycle ranking, indicating a change in which visual tokens receive attention. The denominator of the promotion rate is the second-cycle high-ranked set rather than all visual tokens.

The promoted-token row further distinguishes rank changes inside the annotated target region from those outside it. The two code-reading cases provide an OCR view of the same phenomenon. These selected cases do not estimate how frequently the phenomenon occurs: promotion of individual tokens does not necessarily imply an increase in the total attention assigned to the target region. Therefore, spatial attention reallocation itself should not be equated directly with improved accuracy.

\textbf{Reallocation in natural images.} The three RealWorldQA examples ask about the number of debris pieces beside a white van, the direction of a remote control, and the direction of an arrow on a box. The paired attention heatmaps and promoted-token overlays make changes in question-related spatial reading directly inspectable. The outlined regions are reference annotations rather than regions predicted by the attention mechanism. Together with the color, counting, and code-reading examples above, these natural-image cases illustrate several forms of spatial reallocation. More examples of cross-loop visual reallocation are provided in Appendix~\ref{app:lv-extra-reallocation}.

\textbf{Different reading behavior from a shared attention layer.} In the recorded counting example, the head-averaged attention of the same H-module layer assigns a larger share of visual attention to the annotated target region in the later cycle. Because the same parameters are reused across these invocations, this difference reflects different visual reading behavior of the same shared layer under different recurrent states.

This illustrates how parameter reuse can support different effective reading roles. We do not infer a permanently fixed semantic function for a layer from a single heatmap. The experiment instead provides an aligned example in which, as the recurrent state evolves, the same transformation participates in different visual attention allocations across cycles.

\subsection{Visual Attention Allocation}
\label{sec:lv-5-4}

We distinguish between how much attention is allocated to visual positions and how that attention is distributed within the image. A smaller visual-attention mass can coexist with a more concentrated distribution over visual locations. Neither quantity alone determines whether the attended patches contain the evidence required to answer the question.

We use the visual-token weights \(a_i\) and normalized distribution \(p\) defined in Section~\ref{sec:lv-5-6}. Mass and coverage are computed for each reference-answer query and then averaged equally over queries, heads, and samples.

\textbf{Spatial concentration under shared parameters.} Figure~\ref{fig:lv-lh-comparison}(a) compares normalized spatial entropy at the final layer of each L invocation across the two model cycles. L1, L2, and L3 denote the three invocation positions of the same shared L stack within a cycle. In the second cycle, the mean entropy decreases at all three matched endpoints. This shows that even with exactly the same parameters reused, different recurrent states can produce different visual-attention distributions.

The paired comparison controls the physical layer and invocation position: each L endpoint is compared with the corresponding endpoint in the next model cycle, and the H comparison uses the final layer in both cycles. These are therefore repeated applications of the same transformations to different recurrent states, rather than comparisons between independently parameterized layers. Endpoint comparisons complement the layer-wise trajectories in Section~\ref{sec:lv-5-6} by showing changes between matched points of execution.

For comparison, visual-attention mass is defined as
\begin{equation}
M=\sum_i a_i.
\label{eq:lv-attention-mass}
\end{equation}
Thus, \(M\) measures the total share of attention allocated to visual positions.

Normalization is important for interpreting these measurements. Rescaling all visual attention weights by the same positive factor changes their total mass but leaves the normalized spatial distribution unchanged. A reduction in mass therefore need not imply a broader distribution, and lower entropy need not imply that more total attention is assigned to the image.

Figure~\ref{fig:lv-lh-comparison}(b) aligns the final H layer at the endpoints of the two cycles. Spatial entropy and attention coverage both decrease, while Gini concentration increases. The displayed trajectories follow the same direction, further indicating that visual attention becomes more concentrated at the end of the second cycle.

Spatial entropy and Gini follow the definitions in Equations~\eqref{eq:lv-spatial-entropy} and~\eqref{eq:lv-attention-gini} of Section~\ref{sec:lv-5-6} with their corresponding query scopes. Attention coverage measures the fraction of visual tokens required to accumulate a target share of attention:

\begin{equation}
C_\tau(p)=\frac{k_\tau}{N}.
\label{eq:lv-attention-coverage}
\end{equation}
Here, \(k_\tau\) is the smallest number of highest-weight tokens whose cumulative probability reaches at least the target share \(\tau\). Lower coverage means that fewer visual tokens carry this share of attention.

In Figure~\ref{fig:lv-lh-comparison}, coverage uses a target share of 80\%, making the change interpretable as the fraction of visual positions needed to account for most of the image-directed attention. The measures also differ in their query aggregation. Entropy and coverage summarize reference-answer queries separately before averaging, whereas Gini is computed after pooling instruction and reference-answer attention within each head. Their directional changes describe related but distinct aspects of visual allocation, rather than identical statistics under interchangeable query scopes. The full-set averages characterize this diagnostic collection; they do not establish a universal per-sample trend or identify which concentrated locations provide task-relevant evidence.

\begin{figure}[H]
\centering
\includegraphics[width=\linewidth]{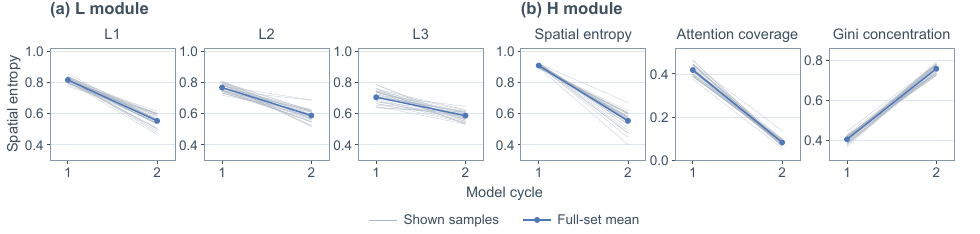}
\caption{Visual-attention concentration under shared-module reuse. (a) Spatial entropy at the final layer of the first, second, and third L invocations (L1--L3), paired across model cycles. (b) Spatial entropy, 80\% attention coverage, and Gini concentration at the final H layer. Gray lines show individual samples. Every blue mean uses all 32 samples. All four entropy panels share a common vertical scale. Entropy and coverage use reference-answer queries; Gini also includes instruction queries. See Appendix~\ref{app:lv-diagnostics} for metric definitions and endpoint indices.}
\label{fig:lv-lh-comparison}
\end{figure}

\textbf{Query/key and value diagnostics.} We further use a cross-combination probe that pairs early and late queries with early and late visual keys. Both early and late query states allocate less attention to late visual keys; within this probe, changing the key state has a larger effect than changing the query state. The same experiment also includes answer-side and instruction-side visual-value removal probes to further examine the role of visual values under different query scopes.

\subsection{State Refinement and Prediction Formation}
\label{sec:lv-5-7}

Attention characterizes how the model reads visual information, while recurrent computation also continuously changes the underlying visual representations. We further analyze how visual states evolve through recurrent computation from three perspectives: hidden-state updates, linear recoverability of initial visual features, and intermediate output distributions.

\textbf{Continued state updates.} Figure~\ref{fig:lv-visual-state-dynamics} measures the changes in visual tokens before and after each complete module invocation. Substantial state updates persist into the second cycle, with comparable mean update magnitudes across the earlier and later halves of execution. The corresponding cosine-similarity matrix further compares representations at the same visual positions across different module endpoints, showing that visual states continue to evolve throughout the recurrent trajectory.

Let \(x_{i,c}\) and \(y_{i,c}\) denote the visual-token vectors before and after module call \(c\), respectively, and let
\begin{equation}
\hat y_{i,c}=\frac{y_{i,c}}{\|y_{i,c}\|_2}
\label{eq:lv-unit-visual-state}
\end{equation}
denote the unit-normalized output. The two panels measure the mean \(L_2\) update magnitude and the cosine similarity between different endpoints:
\begin{align}
U_c&=\frac{1}{N}\sum_i\|y_{i,c}-x_{i,c}\|_2,
\label{eq:lv-state-update}\\
S_{ab}&=\frac{1}{N}\sum_i\hat y_{i,a}^{\top}\hat y_{i,b}.
\label{eq:lv-state-similarity}
\end{align}
Here, \(a\) and \(b\) denote sampled module endpoints. Larger \(U_c\) indicates a more substantial state update, while higher \(S_{ab}\) indicates greater directional similarity between visual representations at the two endpoints. Both statistics are first averaged over corresponding visual positions and then over input samples.

These measures capture complementary aspects of state evolution: update magnitude and endpoint directional similarity. High cosine similarity can coexist with substantial \(L_2\) updates. Together, they describe changes in visual representations, rather than directly measuring how those changes contribute to the final answer.

\begin{figure}[H]
\centering
\includegraphics[width=\linewidth]{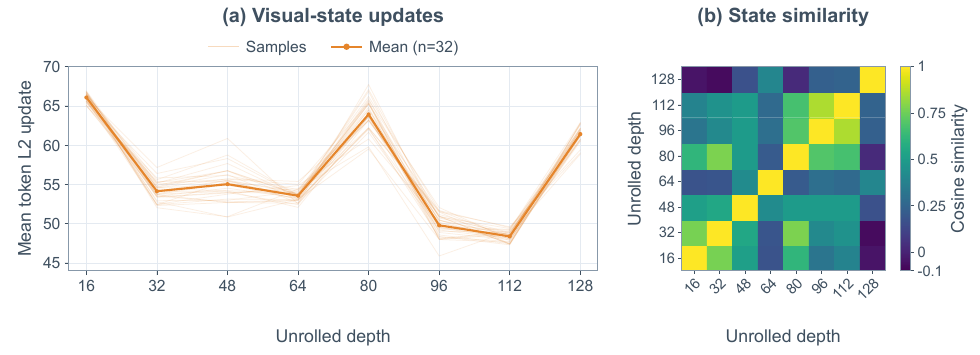}
\caption{\textbf{Visual hidden-state dynamics.} Left: Mean visual-token L2 updates across each complete 16-layer module invocation, with individual sample trajectories and their mean. Right: Cosine similarity between representations at the sampled module endpoints. Continued updates in the second cycle rule out an unchanged-copy description of these states.}
\label{fig:lv-visual-state-dynamics}
\end{figure}

\begin{figure}[H]
\centering
\includegraphics[width=\linewidth]{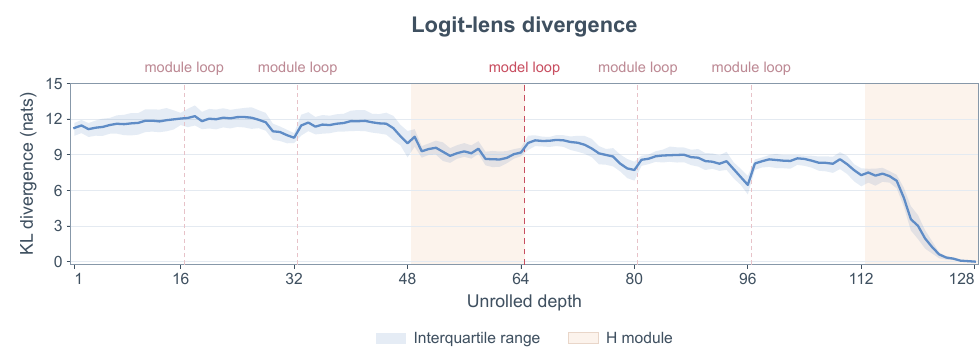}
\caption{\textbf{Logit-lens alignment for the first answer token.} The analysis is based on a 32-sample state-probe collection. The curve shows the mean KL divergence from the final output distribution to each intermediate readout; the blue band represents the sample interquartile range, and pale orange marks H-module invocations. The final H call corresponds to the pronounced late approach of the model output distribution toward the endpoint. Depth counts use one-based indices.}
\label{fig:lv-logit-lens}
\end{figure}

\textbf{Linear recoverability of initial visual features.} Linear-probe analysis shows that the recoverability of the initial visual features changes as recurrent computation proceeds; at the recurrent endpoint, recoverability is lower than immediately after the projector. Recoverability also varies noticeably across module calls, indicating that visual representations continue to be reorganized through repeated applications of the shared modules.

\textbf{Approaching the final model distribution.} Figure~\ref{fig:lv-logit-lens} analyzes the first answer token on the same state-probe collection as Figure~\ref{fig:lv-visual-state-dynamics} and measures the KL divergence between each intermediate readout and the final model output distribution. Divergence remains relatively high during earlier stages, followed by a pronounced decrease within the final H invocation, indicating that the final prediction distribution is formed rapidly during the later stage of recurrent computation.

\begin{samepage}
For each input, let \(P_\ell(v)\) denote the probability of vocabulary token \(v\) at executed layer \(\ell\), obtained using the corresponding module's final normalization and the shared output head. Taking the output distribution at the final depth \(T\) as the reference,
\begin{equation}
K_\ell=\sum_v P_T(v)\log\frac{P_T(v)}{P_\ell(v)}.
\label{eq:lv-logit-lens-kl}
\end{equation}
The sum runs over the entire vocabulary and uses natural logarithms. The curve reports the mean \(K_\ell\) across inputs. Lower KL divergence indicates that the intermediate readout is closer to the final model distribution. The pronounced late decrease in Figure~\ref{fig:lv-logit-lens} shows that the final H module plays a key integrative role in prediction formation.
\end{samepage}

Taken together, these results reveal a clear recurrent visual computation trajectory: visual attention is redistributed across cycles, visual hidden states continue to undergo substantial updates during the second cycle, and the output distribution rapidly approaches the final prediction during the last H computation. Even though the same parameters are shared across cycles, later cycles still perform substantial state transformation and prediction refinement. Repeated application of shared parameters to continuously evolving multimodal states enables deeper vision--language computation.
\endgroup


\section{Related Work}
\label{sec:lv-6}
\begingroup
\clubpenalties 2 10000 0
\widowpenalties 2 10000 0

The literature surrounding LoopVL concerns several related but distinct questions: how a shared transformation is scheduled, what recurrent state carries between invocations, how a visual interface enters that state, and how the resulting computation is evaluated. We organize the discussion around these questions rather than treating every model that repeats an operation as an interchangeable baseline. In particular, the inherited hierarchical language core, the multimodal interface, and the visual diagnostic contribution have different antecedents.

\subsection{Looped Transformers}
\label{sec:lv-6-1}

A looped Transformer reuses an internal transformation during one forward computation. This differs from repeatedly calling an otherwise complete model to generate a sequence of independent textual responses. Universal Transformers provide a shared-depth formulation, while earlier recurrent algorithmic systems and Adaptive Computation Time study repeated updates and the decision to continue computing. Deep equilibrium models provide a related fixed-point perspective, although solving for an equilibrium differs from executing a prescribed finite H/L schedule. These connections establish a lineage of reusable computation without implying that all of the associated training procedures are the same. \citep{dehghani2018universal,kaiser2016neural,graves2016adaptive,bai2019deep}

Later shared-depth systems make the reuse pattern more explicit. MoEUT combines depth-wise parameter sharing with sparse experts, illustrating that recurrent depth need not be a small-capacity dense block. Relaxed Recursive Transformers instead repeat a block of unique layers and add depth-specific low-rank adapters, whereas RingFormer repeatedly applies one Transformer layer with level-specific signals. These designs provide concrete examples of shared computation at different granularities. \citep{csordas2024moeut,bae2025relaxed,heo2025ringformer}

Parameter sharing also appears in models such as ALBERT, while recurrent-memory and block-recurrent Transformers study how state and memory are reused. These designs help distinguish several notions often grouped together: tying parameters across depth, carrying a state between processing steps, and retaining information across input segments. LoopVL's two-cycle visual analysis concerns reuse within its recurrent backbone computation. It does not demonstrate a long-term memory mechanism or persistence across independent image--question episodes. \citep{lan2019albert,bulatov2022rmt,hutchins2022blockrecurrent}

Recent recurrent-depth language models make executed depth an explicit design variable. Huginn and Ouro study repeatedly applied language-model cores, and Loopie describes layer-local recurrence in contrast to model-loop execution. The important point for LoopVL is the granularity of reuse. A complete multi-layer module can be reused without being equivalent to repeating each layer locally, and equal counts of layer applications can produce different execution orders. We use Module-Loop and Model-Loop to make this distinction concrete for the visual--language state. \citep{geiping2025scalinglatent,zhu2025scalinglatent,gao2026loopies}

Executed depth can also be allocated adaptively rather than fixed globally. Mixture-of-Recursions combines a shared stack with token-wise routing over recursion depth, so different token states can receive different amounts of computation. This is complementary to LoopVL's fixed H/L schedule: it broadens the design space, but does not replace controlled comparisons among fixed execution orders. \citep{bae2025mixtureofrecursions}

Our architecture therefore belongs to this broader family of shared-depth models. Its contribution is not that a Transformer can be repeated, nor that stored parameters and executed computation differ. It is the particular multimodal construction and the aligned spatial observations that become possible when image-corresponding positions participate in that computation.

\subsection{Hierarchical and Nested Recurrence}
\label{sec:lv-6-2}

HRM uses coupled high- and low-level recurrent modules, and HRM-Text extends the hierarchical recurrent design to language modeling. We use HRM-Text's open-source framework and data to pretrain our own language model, not its released checkpoint. The L and H state identities, separate module parameters, and nested update schedule remain inherited architectural choices. We describe them because they determine how the multimodal state evolves, not as an independent invention. \citep{wang2025hierarchical,wang2026hrmtext}

Earlier recurrent architectures also imposed structured interactions across timescales. Clockwork RNN partitions state into modules updated at distinct rates, hierarchical multiscale RNNs learn boundary-dependent update, copy, and flush operations, and Nested LSTMs place a recurrent memory inside another recurrent cell. These are antecedents for structured recurrence, but they are not instances of HRM's exact H/L schedule. \citep{koutnik2014clockwork,chung2017hierarchical,moniz2017nested}

The distinction between hierarchical and flat iteration matters even before considering task accuracy. Repeating an operator under one condition and then changing that condition is not generally equivalent to alternating the condition update after every operator call. In LoopVL, the current H state conditions several L updates, and the H update establishes the input context for the next model cycle. This computational distinction motivates same-executed-depth controls, while actual claims about the advantage of one schedule depend on the training and inference protocol.

Other recurrent reasoning systems vary the number of distinct modules and the form of iteration. Tiny Recursive Models provide a compact repeatedly applied network, and analyses of hierarchical models caution that additional choices such as identity conditioning, augmentation, and adaptation can materially affect reported results. These observations motivate separating architectural descriptions from explanations of performance. A table containing different loop counts is informative, but it is not sufficient to isolate every component of a recurrent system. \citep{jolicoeurmartineau2025trm,ge2025hrmperspectives,royeazar2025trm}

The present comparison preserves this distinction. Its inference sweep changes a trained H2L3 model's execution, whereas the training-time ablation independently trains each H/L configuration from initialization and retains that schedule throughout the full training pipeline. A conventional dense language baseline remains a separate control. Similarly, a comparison with a flat recurrent backbone would require its own trained checkpoint rather than relabeling the default architecture. Hierarchical recurrence supplies LoopVL's computational foundation; the visual interface and evidence-allocation diagnostics define the focus of the present work.

\subsection{Visual--Language Interfaces and Spatial Structure}
\label{sec:lv-6-3}

A visual--language interface must connect image features to the representation consumed by a generative language model. \citet{zhou2025perceptioncognition} frame visual alignment as the perceptual foundation for higher-order reasoning. LoopVL uses a pretrained Penguin Vision Encoder and a compact projector, with LLaVA alignment data and LLaVA-OneVision-1.5 multimodal mid-training data in its main stages. This separates visual feature acquisition from adaptation of the language backbone. The model can change the mapping and recurrent interpretation of those features while the encoder parameters remain fixed. \citep{zhang2026penguinvl,liu2023visual,liu2024improved,an2025llavaonevision15}

Prior generative vision--language models illustrate several ways to bridge the two modalities. Flamingo inserts gated cross-attention over a Perceiver-resampled visual stream, whereas BLIP-2 uses a lightweight Querying Transformer between frozen image and language models. LoopVL instead maps retained patch features into language space so that image-corresponding positions can participate directly in recurrent state updates. \citep{alayrac2022flamingo,li2023blip2}

Three interface choices are particularly relevant to recurrence. First, dynamic grids determine how many spatial positions enter the sequence. Second, projector normalization and calibration determine their initial numerical scale. Third, positional coordinates and visibility masks determine how visual and instruction positions interact. Variable-resolution vision has related precedents: NaViT packs images at their native aspect ratios, LLaVA-UHD uses adaptive image slicing, and Qwen2-VL combines native dynamic resolution with multimodal rotary position embeddings. These choices should not be collapsed into a generic stronger-vision explanation. They modify different parts of the computation and require different ablations to establish their effects. \citep{dehghani2023navit,xu2024llavauhd,wang2024qwen2vl}

Projector studies expose a complementary locality--efficiency trade-off. Honeybee reports that generic visual abstractors can weaken local detail when reducing token counts and proposes locality-enhanced projectors, whereas FastV removes visual tokens after using early-layer attention to estimate their importance. LoopVL preserves one projected feature per retained patch rather than adding a token-merging or pruning module, leaving a stable image grid for its spatial diagnostics. \citep{cha2024honeybee,chen2024fastv}

LoopVL's two-axis visual positions are paired with a text mapping compatible with the inherited one-dimensional rotation under the same frequency configuration. PrefixLM visibility allows instruction-conditioned visual updates while preventing answer-to-prefix leakage. The relevant distinction is between making spatial structure available to the recurrent system and proving that the system uses that structure optimally. The architecture establishes the former; the latter requires controlled evidence.

Our later supervised stage combines LLaVA-OneVision-1.5-Instruct-22M, LLaVA-OneVision-2-Spatial-4M, LLaVA, and Vision-R1-Cold, followed by reinforcement learning using Vision-R1-RL. These changes in supervision have a different role from the input interface: a different data mixture is not a new model-loop operation or another image-encoding pass. Likewise, contact-sheet evaluation of multiple images is an input adaptation rather than evidence of native multi-image training. These distinctions keep the multimodal claim tied to the documented implementation. \citep{an2025llavaonevision15,an2026llavaonevision2,liu2024improved,huang2025visionr1}

Recurrent visual reasoning is also studied in LoopViT through visual ARC tasks. Such work is relevant context, but it is not interchangeable with a generative vision--language model using a pretrained language backbone. Our distinction concerns the input, output, and recurrent interface of the task, rather than a claim that visual recurrence itself has no precedent. \citep{shu2026loopvit}

\subsection{Interpreting Recurrent Computation}
\label{sec:lv-6-4}

Alongside work on latent decoding \citep{xu2026lead}, mechanistic studies of recurrent language models examine representation trajectories, repeated inference stages, fixed-point behavior, and the accessibility of information at different depths. These studies discuss mixed, depth-dependent evidence for identifying those trajectories with a literal latent chain of thought. This motivates distinguishing a representation change from an intermediate answer, and an intermediate answer from a causal account of successful reasoning. A probe measures what its chosen readout can access under a particular intervention or normalization, not every computation that the model may implement. Control tasks show that probe accuracy can partly reflect probe capacity, while amnesic probing distinguishes decodable information from information behaviorally used by the model. \citep{blayney2026mechanistic,lu2025latentcot,hewitt2019designing,elazar2021amnesic}

The same distinction is important for attention. A map can show that a query allocates more weight to certain sequence positions. It does not, without additional evidence, establish the complete information transmitted through those positions or their necessity for the final response. Counterfactual attention experiments show that substantially different attention distributions can preserve predictions; subsequent work argues that an interpretability claim must specify what counts as an explanation and use model-aware controls. \citep{jain2019attention,wiegreffe2019attention}

Raw attention is therefore treated here as an allocation diagnostic rather than a standalone causal explanation. V-STAR, for example, reports weaker, more diffuse intermediate-layer visual attention for hallucinated tokens \citep{qian2026cognitivepivot}. Cross-layer mixing and residual paths complicate attribution from one layer's weights alone, and attention distributions need not be uniquely recoverable from a head's output. LoopVL accordingly examines attention amount and conditional spatial concentration separately from query/key compatibility and visual-value perturbations. These measurements address related but nonidentical aspects of the recurrent state. \citep{abnar2020quantifying,brunner2020identifiability}

Vision adds an external spatial correspondence to this analysis. Aligned visual slots can be displayed on the same image while holding a physical module and local layer fixed across invocations. This makes changes in the ranking of locations readily inspectable. It complements text-token visualizations rather than claiming that language representations cannot be visualized. The resulting term visual aha moment describes evidence reallocation in that coordinate system; it does not supply an additional psychological or causal premise.

Our state and readout diagnostics provide complementary views. Substantial late state movement is different from convergence to a final output distribution, and convergence to that distribution is different from correctness against a reference answer. The final-layer KL endpoint is zero by construction because the final model distribution is its own reference. Its preceding trajectory is informative, but a shared output head may have different calibration at intermediate states; work on the tuned lens explicitly learns affine translators to reduce this mismatch. Keeping these distinctions explicit allows the spatial examples, state probes, and task measurements to support one analysis without being treated as interchangeable proof of a single mechanism. \citep{belrose2023tunedlens}

\subsection{Training Regimes and Compute Accounting}
\label{sec:lv-6-5}

Recurrent computation separates the number of stored parameters from the number of transformations executed for a token. This makes comparison criteria especially important. Scaling-law and compute-optimal training discussions distinguish model size, data, and training operations, while recurrent-model studies examine trade-offs among unique depth, loop count, and runtime. Equal parameter storage is therefore not synonymous with equal training FLOPs or equal wall-clock cost. More generally, parameter count, FLOPs, latency, throughput, and memory can disagree, especially when a small set of parameters is reused across many sequential computation steps. \citep{kaplan2020scaling,hoffmann2022training,schwethelm2026much,dehghani2022efficiency}

Deployment-aware scaling analyses further show that the compute-optimal allocation between model size and training data can change once downstream inference demand is included. This reinforces the need to report the boundary of each efficiency claim rather than transferring a training-only comparison to a deployment setting. \citep{sardana2024beyond}

Loopie emphasizes realized training cost and a hardware-aware recurrence recipe. LoopVL does not inherit that study's throughput values, memory results, or compute-matched conclusions. Its own frozen visual encoder, variable sequence lengths, selectively active backward graph, and interface computation require a separate accounting. A parameter-sharing pattern can be described without assuming that a particular implementation realizes the same efficiency advantage on different hardware or under a different optimizer batch schedule. \citep{gao2026loopies}

Training regimes also differ in which positions receive supervision and which parameters are updated. For reasoning supervision, SVSR combines self-correction trajectories with supervised fine-tuning and semi-online preference optimization \citep{qian2026svsr}. LoopVL's main alignment stage trains the visual interface against frozen backbones, while joint training updates our language backbone and the interface. Its later stages change the supervision mixture and training objective. These changes should be specified directly rather than imported through another model's stage name. In particular, a response-only objective does not make the present recipe equivalent to the large-scale supervised pre-training experiment reported for Loopie.

Finally, the data boundary matters as much as the compute boundary. Original records, segmented responses, repeated presentations, valid sequence tokens, and supervised targets describe different aspects of training-data accounting. Studies of data-constrained scaling likewise find that unique data and repeated presentations have different marginal returns. Internal recurrence increases executed computation but not the number of distinct data presentations. LoopVL therefore keeps cumulative-budget claims conditional on the corresponding stage manifests, rather than relying on a small stored parameter count alone. Evaluation comparisons likewise require consistent task variants, prompts, and decoding protocols. \citep{muennighoff2023scaling}

\subsection{Iterative Reasoning and Generalization}
\label{sec:lv-6-6}

A substantial part of the recurrent-model literature studies tasks that can be expressed through repeated updates. Analyses connect looped Transformers to program execution and multi-step optimization, while in-context-learning studies examine whether shared transformations can implement learning procedures. These results motivate reusable computation as an architectural prior. They do not imply that a particular visual attention head performs gradient descent or that every recurrent hidden-state trajectory corresponds to an explicit reasoning algorithm. \citep{giannou2023looped,yang2023looped,gatmiry2024can}

Length and compositional generalization provide a related motivation. On prefix sums, mazes, and chess, recurrent networks trained on easier instances have been shown to solve harder instances by executing additional recurrences at inference time. Excessive iteration can also degrade a solution, a failure mode termed overthinking; recall connections and progressive training have been proposed to stabilize such extrapolation. \citep{schwarzschild2021algorithm,bansal2022end}

Repeated applications of a learned rule may therefore support computations that extend beyond the sequence of transformations observed during training, but the behavior depends on the state representation, step conditioning, and where computation is allocated. These questions are broader than obtaining a higher score by increasing a loop count on one trained checkpoint. \citep{fan2024looped,saunshi2025latentthoughts}

Abstract reasoning tasks make this distinction concrete because a model must infer a transformation or structure from limited task evidence. HRM and related small recurrent systems are relevant examples of this setting. Follow-up analyses emphasize the role of training data, augmentation, adaptation, and evaluation procedure alongside recurrence. LoopVL shares the interest in repeated state refinement but evaluates a different interface and task collection; visual question answering is not assumed to be equivalent to an ARC-style transformation problem. \citep{wang2025hierarchical,jolicoeurmartineau2025trm,ge2025hrmperspectives,royeazar2025trm}

Our inference sweep likewise does not establish length extrapolation, an adaptive stopping rule, or unlimited test-time scaling. It records how a fixed H2L3-trained model responds to alternative execution schedules. The spatial diagnostics may suggest mechanisms worth testing in language-only recurrence, but they do not demonstrate that those models use the same mechanism. Connecting cross-loop visual reallocation to systematic generalization would require additional controlled evaluations. This separates the broader motivation for recurrent reasoning from the narrower observations currently available for LoopVL.
\endgroup


\section{Future Work}
\label{sec:lv-7}

Our reasoning-oriented post-training experiments focus primarily on mathematical reasoning. Under the current limited-data setting, extended reasoning can occasionally suffer from repetitive tokens or repeated reasoning segments, which may interrupt the reasoning process and prevent the model from completing a valid solution. A natural direction for future work is to continue post-training LoopVL on a broader range of tasks, including scientific reasoning, instruction following, and agentic task-solving, while substantially increasing both the volume and diversity of post-training data. Incorporating richer preference and reasoning supervision may further improve generation stability, reasoning reliability, and the model's general-purpose capabilities.

Second, due to computational constraints, we have not trained a separate vision--language model with a more general Model-Loop architecture as a matched control for studying Visual Aha Moments. Although our current experiments show that the observed cross-cycle attention shift persists under different backpropagation routes and after removing the visual anchor--gate mechanism, they do not fully establish whether Visual Aha Moments are intrinsically associated with Model-Loop computation or instead depend on other architectural or training choices specific to LoopVL. We are preparing to train the next generation of LoopVL and plan to conduct broader experiments to examine whether this phenomenon persists at larger model scales and in more general Model-Loop architectures.

Third, LoopVL adopts a hierarchical recurrent language architecture built upon the HRM-Text framework, while the primary focus of this work is to investigate how visual tokens evolve when repeatedly processed by a Loop Transformer. Our current results therefore characterize recurrent visual computation under one particular hierarchical framework rather than exhaustively exploring the broader design space of recurrent architectures. As LoopVL scales to larger models, selecting an appropriate recurrent framework that can scale effectively while naturally accommodating visual-token computation remains an important direction for future work. We plan to investigate alternative recurrent organizations, variable-depth computation, adaptive recurrent schedules, and inference-compute-matched settings to better understand how the choice of recurrent architecture affects multimodal reasoning, scalability, and computational efficiency.

Finally, our current study primarily considers image-based vision--language tasks with a frozen visual encoder. The recurrent backbone repeatedly refines visual states derived from a fixed set of encoded image features, leaving several broader multimodal settings unexplored. Extending recurrent visual computation to native multi-image understanding, video, and interactive or agentic environments may require recurrent states to integrate visual evidence not only across computational depth but also across images and time. Future work will explore these settings, as well as joint adaptation of the visual encoder and recurrent backbone, to investigate how visual representations and recurrent computation can be co-designed for larger and more general multimodal models.

\section{Conclusion}
\label{sec:lv-8}

We introduced LoopVL, a vision--language model whose language backbone we pretrain using HRM-Text's open-source framework and data, with nested Module-Loop and Model-Loop computation. Its visual interface combines dynamic grids, calibrated embeddings, and compatible spatial positions. The current analysis distinguishes schedule sensitivity, attention concentration, cross-loop relevance reallocation, state updates, and final-readout alignment. These observations motivate visual aha moments as a spatially interpretable description of recurrent evidence use, while leaving stronger causal and generalization claims to controlled evaluation. LoopVL thus provides a framework for examining what changes when shared modules process visual evidence again.

\section{Acknowledgments}

\begin{itemize}
    \item We thank Prof. Zhuoran Lu and Prof. Yanzhe Zhang at The University of Hong Kong for their valuable feedback and careful review.

    \item We thank the authors of Loop the Loopies! \citep{gao2026loopies} for open-sourcing their elegant \LaTeX{} template used in this work.

    \item We thank the authors of HRM-Text \citep{wang2026hrmtext} for open-sourcing their recurrent language modeling framework, which provided an important foundation for this work.
\end{itemize}

\LVAuthorCheck{AUTHOR_METADATA}{Author names, contributions, acknowledgments, and public model/code links await author confirmation.}

\bibliographystyle{abbrvnat}
\bibliography{loopvl_references}

\clearpage
\appendix
\phantomsection
\addcontentsline{toc}{section}{Appendix}
\addtocontents{toc}{\protect\loopvlappendixtocstyle}
\makeatletter
\renewcommand{\toclevel@section}{2}
\renewcommand{\toclevel@subsection}{3}
\renewcommand{\toclevel@subsubsection}{4}
\makeatother
\section{Evaluation Settings and Scoring Methods}
\label{app:lv-evaluation}

We evaluate LoopVL using a unified inference and scoring pipeline, with VLMEvalKit providing dataset loading and prompt construction for a subset of the benchmarks. Evaluation primarily uses direct-answer prompts, greedy decoding, and a batch size of 1, with the random seed set to 0. The generation budget is 32 tokens for most examples, with fixed budgets of 8, 64, or 512 tokens for selected examples. Predictions are scored using option extraction, answer normalization, and task-specific matching rules.

ChartQA uses the relaxed accuracy implementation from VLMEvalKit, allowing a 5\% relative error for numerical answers. For POPE, we report F1 separately for the adversarial, popular, and random categories, and compute pooled F1 from the TP, FP, and FN counts aggregated across category instances. HallusionBench scores are the arithmetic mean of aAcc, fAcc, and qAcc. For MMEval-Pro, a triplet is counted as correct when its Origin, Perception, and Knowledge questions are all answered correctly; the final score is the equally weighted macro-average of genuine accuracy across the three source datasets. MMMU-Pro uses a fixed evaluation set of 1,730 questions.

The main model uses the H2L3 configuration, following the execution order \texttt{LLLHLLLH}. Each module contains 16 layers, yielding 128 layer executions per forward pass. The H/L sweep varies the recurrence schedule of the same checkpoint across 12 configurations, with H ranging from 1 to 4 and L from 1 to 3. The H state is initialized from the multimodal input embeddings, and the L state uses the initial state stored in the checkpoint. Decoding runs with the KV cache disabled, and the final H-module output is passed through the shared output head to generate answers. All sweep configurations use a 32-token generation budget.

\section{Diagnostic Metrics and Sample Sources}
\label{app:lv-diagnostics}

The attention curves and L/H comparisons use the same set of 32 examples, comprising four visual question-answering task families with eight examples each. Spatial entropy and attention coverage are computed from reference-answer queries on each query's visual attention distribution, then averaged over queries, the 12 attention heads, and examples. Coverage measures the fraction of visual tokens required to accumulate 80\% of the attention mass. Gini is computed by first aggregating raw attention from instruction and reference-answer queries within each head, followed by normalization over visual tokens and concentration measurement. The query-scope analysis evaluates different query subsets using the same teacher-forced forward pass.

Case-study heatmaps average reference-answer attention and normalize it over the complete visual grid. Promoted tokens are visual tokens whose attention values fall at or below the 50th percentile in the first cycle and at or above the 75th percentile in the second cycle. The promotion rate is the proportion of promoted tokens among the high-attention tokens in the second cycle. Target regions are defined by spatial annotations, with manually annotated bounding boxes or polygons used for real images.

The state and output probes use 16 evenly spaced examples from VMCBench and 16 from AI2D, with images and question prompts as inputs. Visual L2 update measures representation changes across a complete 16-layer module invocation, while cosine similarity compares representations at corresponding visual positions. The logit lens applies the final normalization of the corresponding stack and the shared output head to predict the first answer token. KL divergence uses the final prediction distribution as the reference and is computed with natural logarithms. Shaded bands indicate the interquartile range across examples. Attention indices range from 0 to 127, while state and output probes use cumulative layer-execution counts from 1 to 128. Full-depth attention curves use PCHIP interpolation; paired comparisons and logit-lens curves use the original recorded values.

\clearpage
\section{Additional Visual Reallocation Examples}
\label{app:lv-extra-reallocation}

\begin{figure}[!htbp]
\centering
\includegraphics[width=\linewidth]{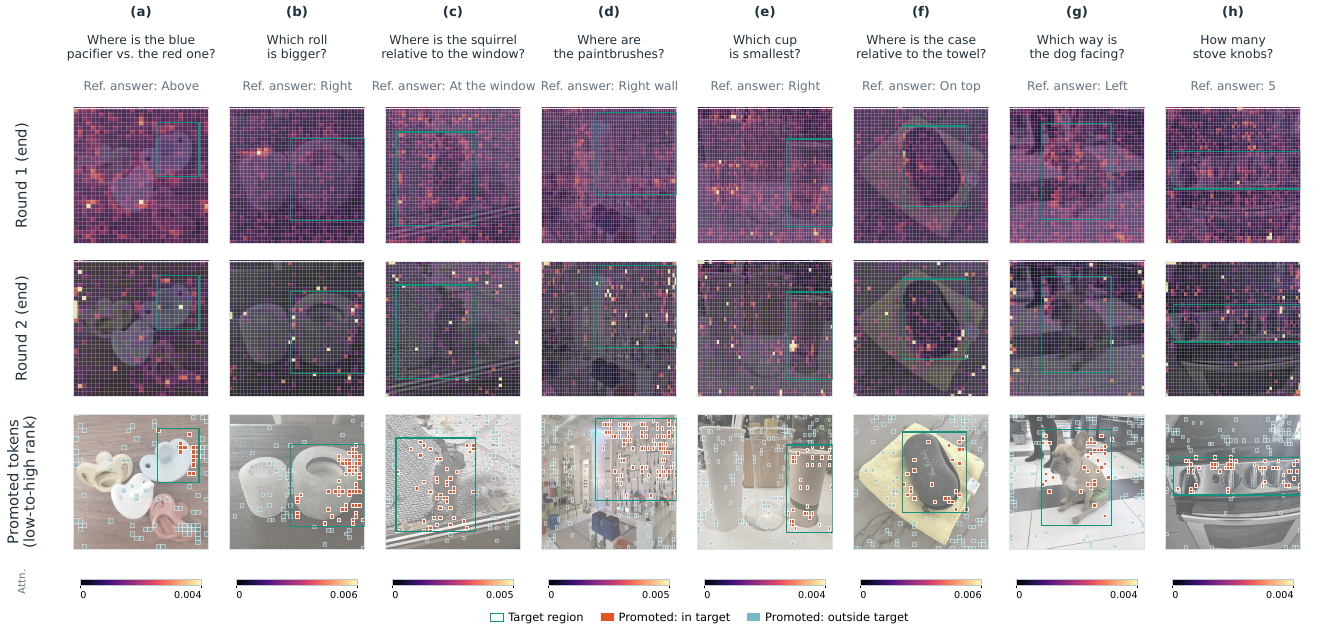}
\caption{Additional visual reallocation examples, Set 1. Rows compare the same final H-module layer at the two cycle endpoints and mark tokens promoted from the first-cycle bottom half to the second-cycle top quartile. Orange and cyan mark promoted tokens inside and outside the outlined reference regions, respectively. Attention is normalized over the full visual grid separately in each cycle; paired heatmaps share a color scale. Images and overlays are jointly resized for display only. Labels give reference answers, not intermediate predictions.}
\label{fig:lv-extra-reallocation-1}
\end{figure}

\begin{figure}[!htbp]
\centering
\includegraphics[width=\linewidth]{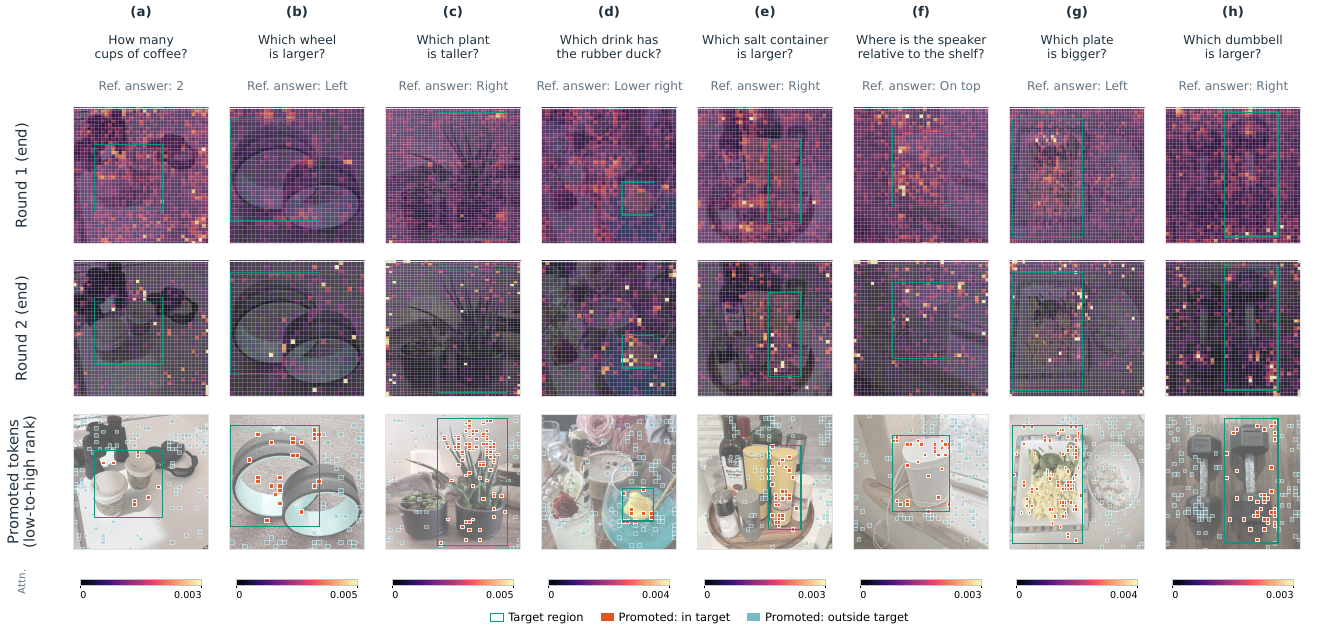}
\caption{{\bfseries Additional visual reallocation examples, Set 2.} Eight additional counting, size-comparison, and spatial-relation examples follow the same row layout, normalization, and promotion rule as Figure~\ref{fig:lv-extra-reallocation-1}. Rank promotion of individual tokens does not necessarily imply an increase in the total attention within the annotated target region.}
\label{fig:lv-extra-reallocation-2}
\end{figure}

\begin{figure}[!htbp]
\centering
\includegraphics[width=\linewidth]{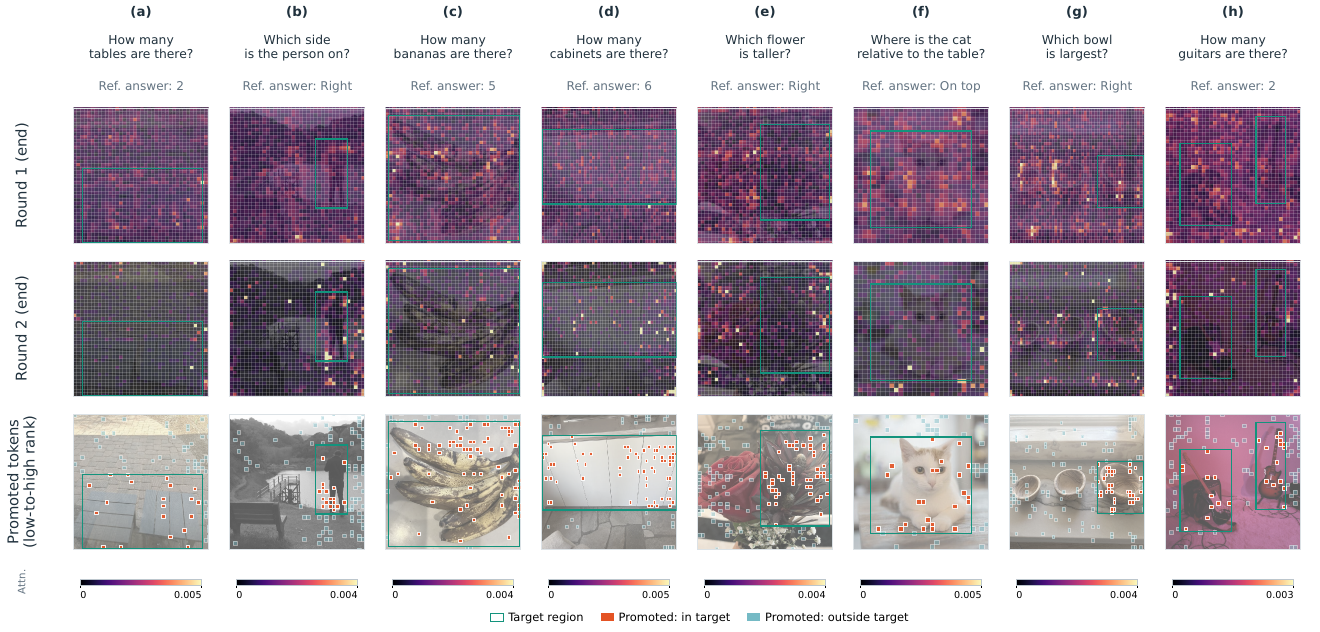}
\caption{{\bfseries Additional visual reallocation examples, Set 3.} Eight additional counting, size-comparison, and spatial-relation examples follow the same row layout, normalization, and promotion rule as Figure~\ref{fig:lv-extra-reallocation-1}. Rank promotion of individual tokens does not necessarily imply an increase in the total attention within the annotated target region.}
\label{fig:lv-extra-reallocation-3}
\end{figure}

\begin{figure}[!htbp]
\centering
\includegraphics[width=\linewidth]{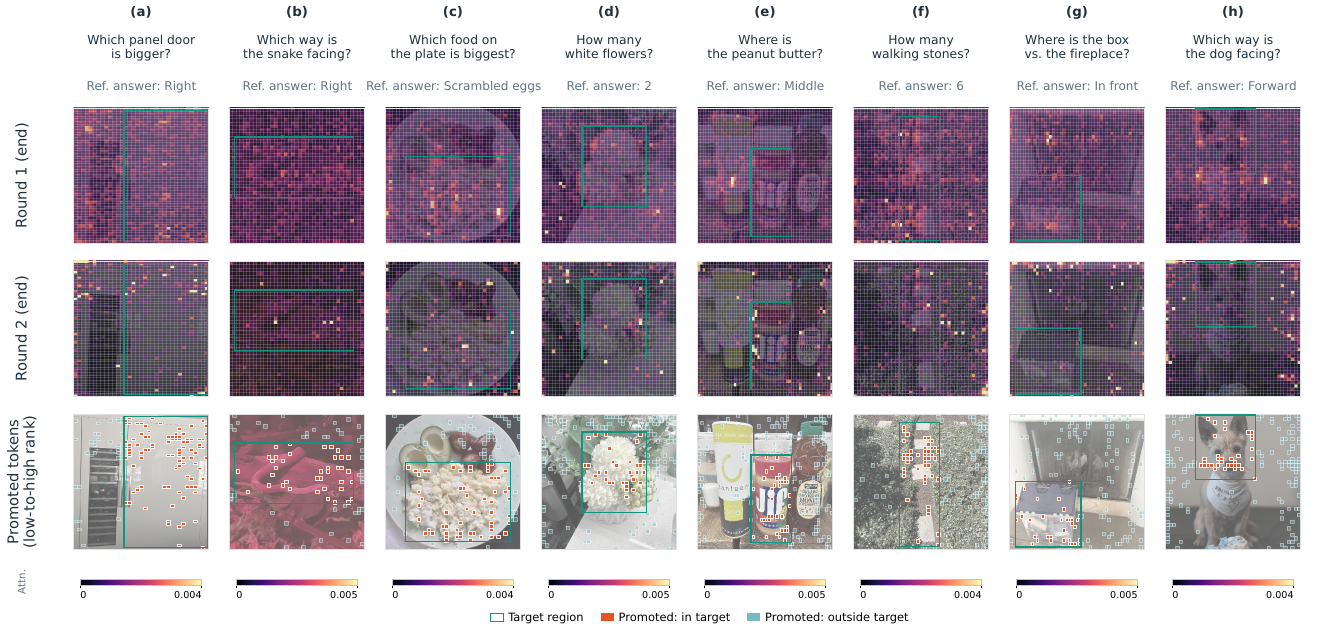}
\caption{{\bfseries Additional visual reallocation examples, Set 4.} Eight additional counting, size-comparison, and spatial-relation examples follow the same row layout, normalization, and promotion rule as Figure~\ref{fig:lv-extra-reallocation-1}. Rank promotion of individual tokens does not necessarily imply an increase in the total attention within the annotated target region.}
\label{fig:lv-extra-reallocation-4}
\end{figure}

\begin{figure}[!htbp]
\centering
\includegraphics[width=\linewidth]{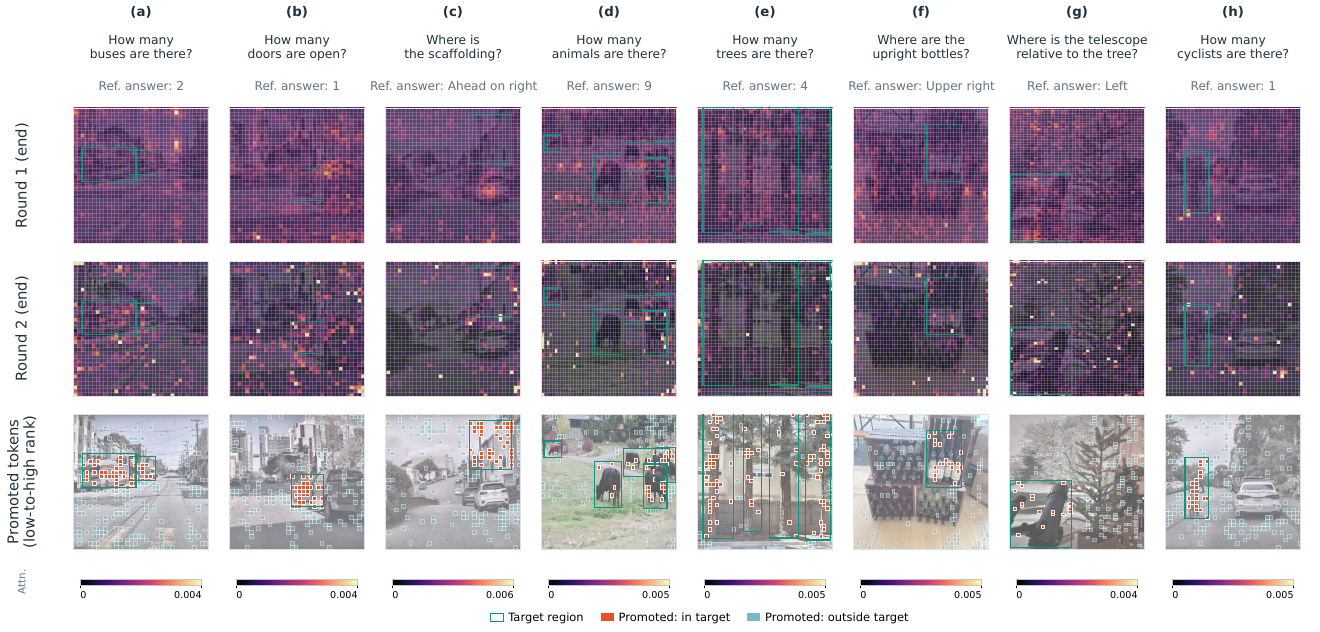}
\caption{{\bfseries Additional visual reallocation examples, Set 5.} Eight additional counting, size-comparison, and spatial-relation examples follow the same row layout, normalization, and promotion rule as Figure~\ref{fig:lv-extra-reallocation-1}. Rank promotion of individual tokens does not necessarily imply an increase in the total attention within the annotated target region.}
\label{fig:lv-extra-reallocation-5}
\end{figure}

Figures~\ref{fig:lv-extra-reallocation-1}--\ref{fig:lv-extra-reallocation-5} extend Figure~\ref{fig:lv-reallocation-cases} with 40 additional selected RealWorldQA examples in token-grid form. Attention maps use reference-answer queries averaged over heads, not free-generation queries. These qualitative cases illustrate spatial reallocation, without estimating its prevalence or establishing a causal improvement in answer accuracy.

\clearpage

\end{document}